\documentclass[journal]{IEEEtran}

\usepackage{algorithm}
\usepackage{array}
\usepackage[caption=false,font=normalsize,labelfont=sf,textfont=sf]{subfig}
\usepackage{textcomp}
\usepackage{stfloats}
\usepackage{url}
\usepackage{verbatim}
\usepackage{graphicx}
\usepackage{cite}
\usepackage{booktabs}
\usepackage{algpseudocode}
\usepackage{multirow}
\usepackage[T1]{fontenc}
\usepackage{hyperref}
\usepackage{courier}
\usepackage{listings}
\usepackage{tcolorbox}
\usepackage{amsmath, amssymb, amsfonts, mathtools}
\usepackage[colorinlistoftodos]{todonotes}
\usepackage{subcaption} 
\usepackage{kotex}
\usepackage{cleveref}
\usepackage{graphicx}    
\usepackage[table]{xcolor}   
\usepackage{array}           
\usepackage{arydshln}
\usepackage[switch]{lineno}

\definecolor{darkgreen}{rgb}{0.0, 0.5, 0.0}
\tcbuselibrary{listings,breakable}
\begin{document}

\newcommand{\rev}[1]{\textcolor{red}{#1}}
\newcommand{\mrev}[1]{\textcolor{blue}{#1}}

\title{PCGRLLM: Large Language Model-Driven Reward Design for Procedural Content Generation Reinforcement Learning}

\newcommand{\orcidlogo}{\includegraphics[height=9pt]{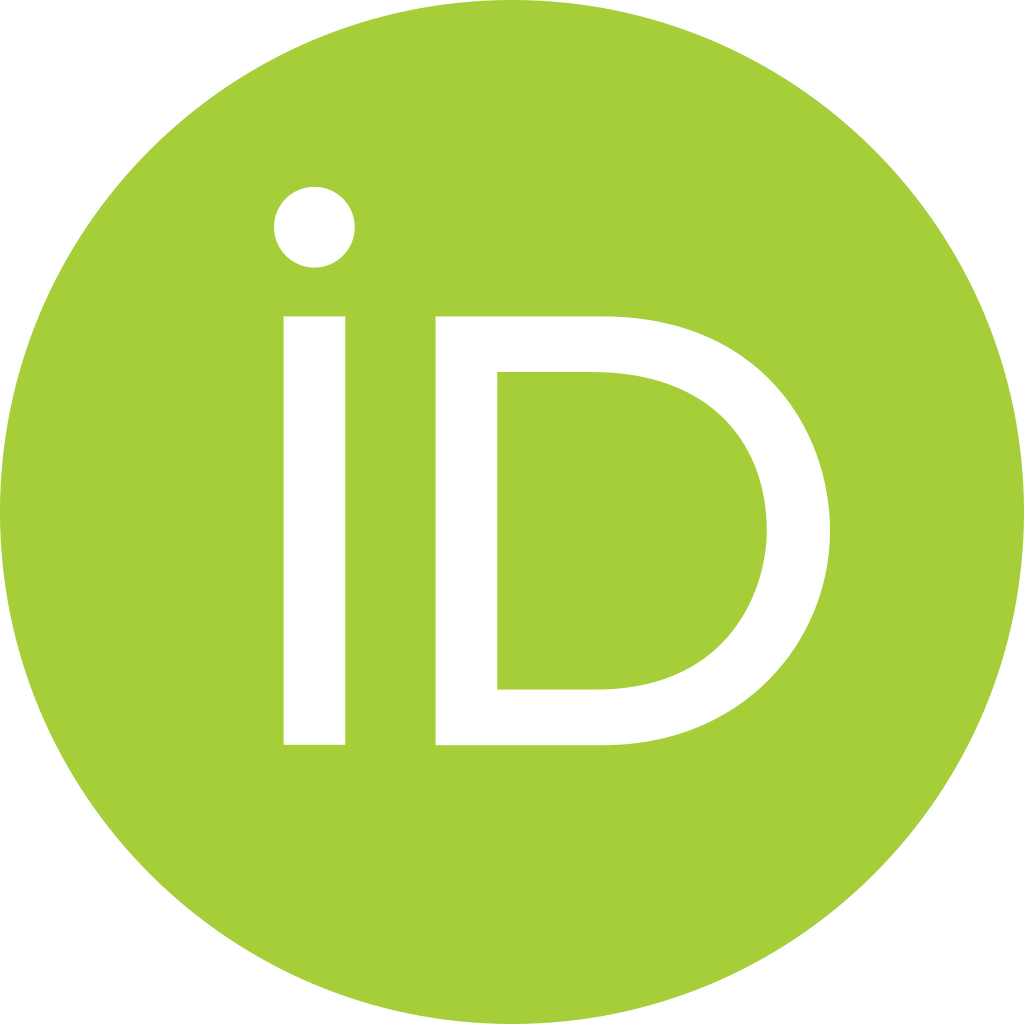}}

\newcommand{\orcidlink}[1]{\href{https://orcid.org/#1}{\orcidlogo}}

\author{
In-Chang Baek\IEEEauthorrefmark{2},
Sung-Hyun Kim\IEEEauthorrefmark{2},
Sam Earle\IEEEauthorrefmark{3},
Zehua Jiang\IEEEauthorrefmark{3},
Jin-Ha Noh\IEEEauthorrefmark{2},
Julian Togelius\IEEEauthorrefmark{3},
Kyung-Joong Kim\IEEEauthorrefmark{2}\IEEEauthorrefmark{4}
\\
\IEEEauthorrefmark{2}Gwangju Institute of Science and Technology (GIST), Gwangju, South Korea \\
\IEEEauthorrefmark{3}New York University, New York, USA \\
inchang.baek@gm.gist.ac.kr, kjkim@gist.ac.kr
\thanks{\IEEEauthorrefmark{4}Corresponding author}

}


\maketitle

\begin{abstract}
Reward design plays a pivotal role in the training of game AIs, requiring substantial domain-specific knowledge and human effort.
In recent years, several studies have explored reward generation for training game agents and controlling robots using large language models (LLMs).
In the content generation literature, there has been early work on generating reward functions for reinforcement learning agent generators.
This work introduces \textit{PCGRLLM}, an extended architecture based on earlier work, which employs a feedback mechanism and several reasoning-based prompt engineering techniques.
We evaluate the proposed method on a story-to-reward generation task in a two-dimensional environment using two state-of-the-art LLMs across various reasoning-based prompting methods.
Our experiments provide insightful evaluations that demonstrate the capabilities of LLMs essential for content generation tasks.
The results demonstrate a substantial performance improvement over the previous structure, achieving performance comparable to that of humans.
Our work demonstrates the potential to reduce human dependency in game AI development, while supporting and enhancing creative processes.

\end{abstract}


\section{Introduction}
The design of reward functions plays a pivotal role in the training of deep reinforcement learning (DRL) agents and evolutionary algorithms for procedural content generation (PCG) in games \cite{liu2021deep}. Reward functions guide agent behaviors and define the objectives that align generated content with desired outcomes, such as game difficulty or aesthetic appearance.
Traditionally, designing reward functions relies heavily on researchers' game-specific knowledge and time-consuming reward shaping process. 
In the procedural content generation via RL (PCGRL) literature \cite{khalifa2020pcgrl}, the controllability of reward function has been achieved by parameterization of reward function in two- and three- dimensional level generation tasks \cite{earle2021learning, jiang2022learning}.
This significant human dependency not only requires significant time and resources but also introduces barriers to accessibility and scalability of game AIs.
Additionally, the controllability of RL-based generative models has been dependent on pre-defined environmental features.
Therefore, reward generation is necessary to alleviate the dependency on humans and dependency on controllable features.

\begin{figure}[!t]
    \centering
    \includegraphics[width=\linewidth]{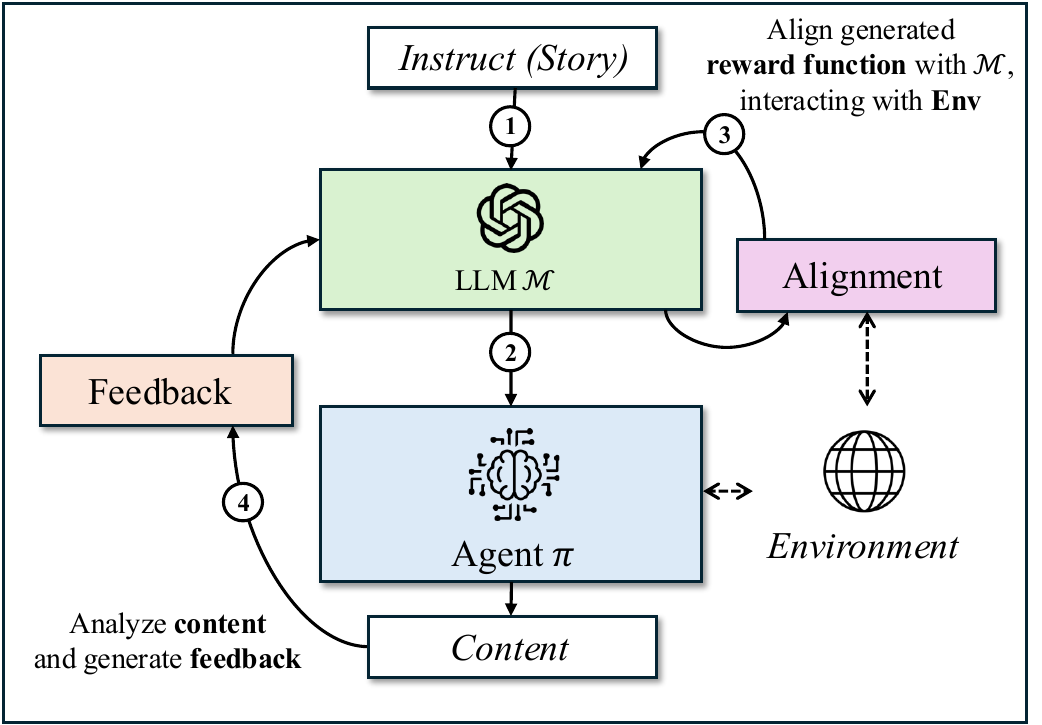}
    \caption{An overview of the reward generation process: (1) instructions guide the LLM, (2) outputs direct the agent, (3) environment interactions refine rewards, and (4) feedback analyzes content for improvement.}
    \label{fig:concept}
\end{figure}

Recent advancements in LLMs have shown their potential to mitigate these challenges by leveraging pre-trained expert knowledge from large datasets. Several studies have explored LLM-based reward generation approaches in robotic control \cite{ma2023eureka, zeng2023learning, yu2023language} and the gameplay \cite{li2024auto, zheng2024online} domain, utilizing LLMs' reasoning and coding capabilities.
One such approach, \textit{ChatPCG} \cite{baek2024chatpcg}, introduced an early-stage LLM-driven reward generation method for PCGRL, which transforms high-level game descriptions into low-level reward function code.
This work proposed a reward fine-tuning method, self-alignment, to align the reward function for a specific environment. However, a limitation of this approach is the absence of a refinement process that incorporates results of the trained agent. As a result, it is uncertain whether the trained policy accurately reflects the intended reward function generation conditions.

To address these limitations, we propose an improved architecture based on prior work \cite{baek2024chatpcg}, \textit{PCGRLLM}, a feedback-based reward generation framework for content generation.
Fig. \ref{fig:concept} illustrates the overview of the proposed method with the sequence of improving reward function with two major refining processes: self-alignment and feedback.
The language model generates a reward function with a brief story instruction and a PCGRL model is trained with the generated function.
To ensure that the RL agent generates content that satisfies the instruction, the language model provides feedback and updates the reward function in the next iteration.
The feedback allows the RL agent to improve its policy by observing the actual outcomes of the trained agent, while the LLM generates rewards that can be effectively incorporated into the RL agent's training.

Specifically, our contributions are threefold:
\begin{itemize}
\item \textbf{Enhancing reward generation architecture:} The feedback mechanism enhances the reward generation pipeline, enabling the policy to align more effectively with the given instructions.
\item \textbf{Reasoning-based refinement prompting:} State-of-the-art prompt engineering techniques are employed to improve the exploration of the reward space.
\item \textbf{Comprehensive modular evaluation:} Extensive experiments provide ablation study and extensive insights into the capabilities of LLMs in content generation.
\end{itemize}

We evaluate the proposed framework with a text-to-reward task, which evaluates reward functions based on how they reflect a given textual prompt.
The PCGRL agent is trained with LLM-generated reward functions, and the quality of these reward functions is evaluated based on the agent-generated content.
The task is demonstrated in a two-dimensional level generation environment \cite{khalifa2020pcgrl,earle2024scaling}, with a brief story input, such as: \textit{"The player needs to obtain a key and escape through the door. To pick up the key, the player encounters bat monsters."}
To generate reward values for content generation, it is essential to design reward functions iteratively through effective planning and to employ reasoning to assess the causal relationship between the reward function and the resulting content.
From the perspective of content generation, we evaluate the reasoning capabilities of LLMs---code  generation, content evaluation, reflection, and key extraction---by conducting extensive evaluations.

\section{Background}

\begin{figure*}[!t]
    \centering
    \includegraphics[width=\linewidth]{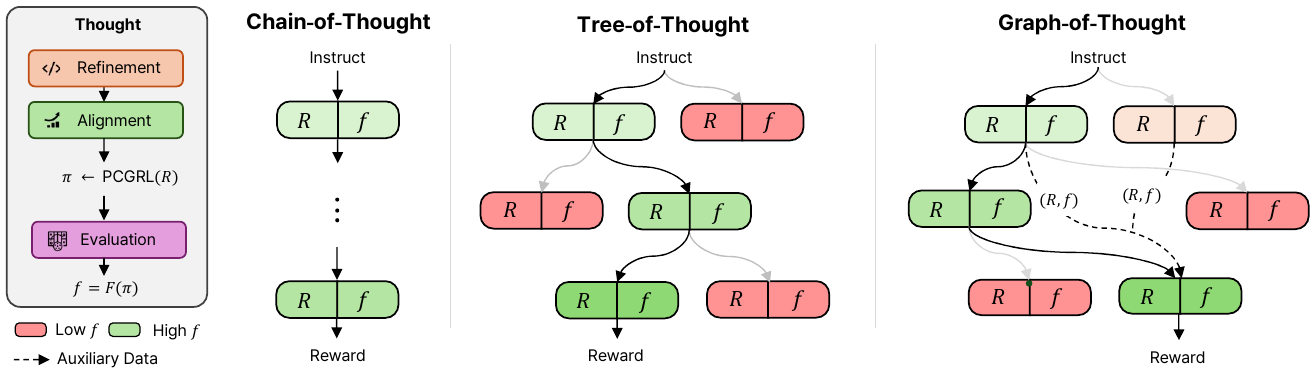}
    \caption{The architectural comparison of three prompt engineering techniques, along with details of the thought nodes. Each thought node includes a reward function ($R$) and a fitness value ($f$), which represent the evaluated score of the contents trained by the agent using the reward function. In the Tree- and Graph-of-Thought methods, the parent node is selected based on the fitness value.}
    \label{fig:reasoning_pe}
\end{figure*}

\subsection{Reward Generation Problem}

The reward design problem \cite{lewis2010rewards} refers to the process of searching for a reward function ($R$) that maximizes the fitness score $F(\pi_{\text{trained}})$, where $\pi_{\text{trained}}$ represents a policy trained using $R$.
A DRL agent takes an action $a$ in a state $s\in{}S$ and receives a reward value $r=R(s|a)$.
The reward function can be designed as a sparse reward based on task success criteria or as an immediate dense reward for actions.
Recently, there has been significant interest in LLMs for reward generation, particularly in the robotics domain, where these models benefit from extensive human-level prior knowledge. Compared to using LLMs as policies, LLM-based reward generation is advantageous for reducing inference costs of large models and improving performance in low-level control tasks \cite{ahn2022can, singh2023progprompt}.

For instance, studies such as \cite{yu2023language, zeng2023learning, song2023self} have demonstrated the use of LLMs to generate reward functions for training DRL agents in robotics. By providing detailed environment descriptions and task-specific rules, these approaches facilitate reward generation for robotic tasks. Building on these efforts, \textit{Eureka} \cite{ma2023eureka} and \textit{Text2Reward} \cite{xie2023text2reward} introduced advanced techniques such as evolutionary search and reward reflection to achieve human-level reward design performance. These methods also incorporate mechanisms to evolve reward functions using human feedback or preferences.
Notably, LLMs take a leading role in reward function generation, either fully automating the process or assisting human creativity in designing effective rewards.
While most of these studies focus on robot control tasks by generating dense rewards for solving complex tasks, there remains a need to explore their potential applications in domains like content generation.

\subsection{Procedural Content Generation via RL}
PCGRL \cite{khalifa2020pcgrl}---a DRL-based content generation method---is a machine learning-based content generation method.
The generator agent is trained with a hand-crafted reward function and gets a positive reward when the content gets closer to the goal condition.
The benefits of PCGRL stem from its data-free nature and computational efficiency during inference, making it well suited for real-time content generation in games \cite{togelius2011search}.
PCGRL was first applied to environments such as mazes, \textit{Zelda}, and \textit{Sokoban} \cite{khalifa2020pcgrl}.
More recent work has extended this line of research by focusing on \textit{Sokoban} with deep learning–based approaches, incorporating methods like diversity sampling to improve solution variety \cite{zakaria2022procedural}.
Reward design is a longstanding challenge in reinforcement learning broadly. It remains particularly important in PCGRL, where the diversity and quality of generated levels largely depend on the reward functions.
These advancements include support for conditional metric inputs \cite{earle2021learning}, language-conditioned policy \cite{baek2025ipcgrl,kim2026multi,baek2025human}, the ability to freeze specific tiles during generation \cite{earle2024scaling}, competitive level balancing \cite{rupp2024simulation}, 3D level domain \cite{jiang2022learning}, balancing  multiplayer game skill data \cite{jeon2023raidenv}, and graph representation for game economics and skill trees \cite{rupp2024g}.

In PCGRL, the level design process is framed as a Markov Decision Process (MDP), where level generation is learned through a trial-and-error approach. At each step \( t \), the agent observes the game level as a state \( s_t \), selects an action \( a_t \) to modify a tile of the level, and transitions to a new state \( s_{t+1} \).
The agent then receives a reward: $r_t = R(s_t, s_{t+1})$, determined by a reward function ($R$) that evaluates the transition between states.
In PCGRL, reward function design requires identifying a computable target based on the given generation objective and appropriately combining different weights.
When multiple sub-functions need to be combined to achieve the desired artifact, designing a reward function in a single attempt is highly challenging. Instead, the reward functions has been refined by humans through multiple attempts, with iterative modifications made based on the observed results.
Therefore, the reward design process involves iteratively combining functions to generate a reward function that produces game-like levels while satisfying the given conditions.
The environment details are described in Appendix~\hyperlink{sec:environment_details}{A}.

\subsection{LLM-driven Reward Generation for Procedural Content Generation}

Reward design has been a crucial aspect of content generation, but it remains heavily reliant on expert knowledge and is often time-consuming. Traditionally, reward functions have been either fixed through manually crafted code \cite{khalifa2020pcgrl} or conditionally generated based on predefined features, such as path length, determined by experts \cite{earle2021learning}. This dependency on predefined conditions limits the diversity of content that can be generated and constrains the flexibility of reward design. Furthermore, the need for domain-specific knowledge about the game reduces the accessibility of content generation algorithms, making them less adaptable to diverse applications.

\textit{ChatPCG} \cite{baek2024chatpcg} introduced an early architecture that employs LLMs to automatically generate and refine reward functions for PCGRL.
ChatPCG refines reward functions with game-specific variable magnitudes using a self-alignment technique. 
However, it lacked feedback from the actual trained policies, making it difficult to predict content outcomes from arbitrary functions. More specifically, the prior approach did not evaluate the behaviors of trained RL policies as a source of feedback, and systematic approaches to overcoming this limitation have yet to be explored.

To train a DRL agent, the reward function serves as the primary learning signal, shaping credit assignment, training stability, and final performance.
While inference is fast once a policy is trained, the training process is computationally expensive, making reward validation and refinement especially crucial.
In this context, it is important to note that both pre-training and post-training phases of policy learning can benefit from LLM-based tuning, which enables iterative refinement throughout the entire process.
For these reasons, this paper primarily focuses on RL-based content generation, with an emphasis on automated reward design and self-feedback mechanisms to improve efficiency.

\subsection{Prompt Engineering}
Prompt engineering (PE) has become an effective methodology to enhance the performance of large language models (LLMs) in a gradient-free manner, with various techniques developed to improve their logical reasoning and planning abilities. These techniques can be described as utilizing different structures to extend thought processes, as illustrated in Fig. \ref{fig:reasoning_pe}. Compared to traditional input-output (IO) methods that query results in a single step, these structured approaches show superior performance in solving problems sequentially.

These techniques can be broadly categorized into three approaches based on chain, tree, and graph structures:

\textbf{Chain-of-Thought (CoT)} \cite{wei2022chain,kojima2022large} expands reasoning process with multiple steps to solve a problem in step-by-step. This method is particularly effective for problems requiring long-horizon reasoning. An enhanced variant, CoT with self-consistency (CoT-SC) \cite{wang2022self}, selects the most consistent response through a majority vote mechanism.

\textbf{Tree-of-Thoughts (ToT)} \cite{yao2024tree} expands reasoning in multiple directions, increasing the scope of exploration.
This fault-tolerant method, equipped with backtracking capability, not only broadens the exploration space but also allows recovery by redirecting to alternative paths when a wrong direction is taken, providing resilience against getting trapped in local optima.
The multi-path reasoning requires a fitness function to evaluate and select nodes for expansion, ensuring efficient exploration of the solution space.

\textbf{Graph-of-Thoughts (GoT)} \cite{besta2024graph} is an extended version of ToT, designed to improve the sample efficiency of reasoning by leveraging multiple thoughts generated during node expansion. When expanding a node, GoT retrives related nodes to enhance the reasoning process and ensure efficient exploration of the solution space.

\begin{figure*}[th]
    \centering
    \includegraphics[width=1.0\linewidth]{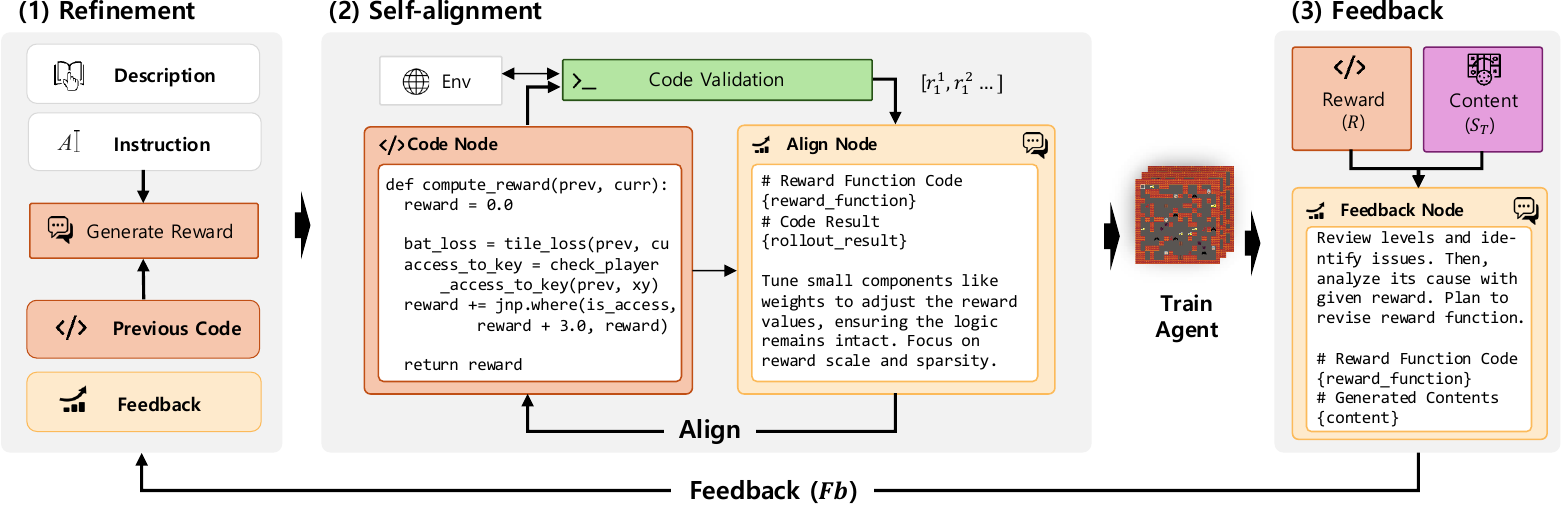}
    \caption{Architecture of PCGRLLM framework. "Message icons \includegraphics[height=0.8em]{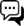}" indicate the use of language model ($\mathcal{M}$) in the context.}
    \label{fig:architecture}
    \vspace{-0.34cm}
\end{figure*}

\section{Story-based Reward Function Generation Task}
\label{sec:story2reward}
Recent advancements in text-based generative models have showcased the potential for translating textual descriptions into diverse domains such as human-like motion \cite{guo2022generating}, high-fidelity images \cite{ramesh2022hierarchical}, music composition \cite{agostinelli2023musiclm}, and game content generation.
The gaming domain has also benefited significantly from text-based generative approaches. For example, text-conditioned generative models have been applied to specific tasks such as generating \textit{Super Mario Bros} levels \cite{sudhakaran2024mariogpt} or \textit{Sokoban} puzzles \cite{todd2023level}, where models synthesize playable and contextually relevant game content.
Extending beyond level design, recent works have explored generating entire games from textual descriptions \cite{zhang2024text, nasir2024word2world}, thereby transforming abstract narratives into interactive environments and mechanics.
The generated content is evaluated to ensure it aligns with the instructions (i.e., textual conditioning).

Our method is evaluated on the text-to-reward generation task, which aims to bridge narrative-driven descriptions with a trainable reward function. This evaluation checks whether the generated content satisfies the given text instructions, such as ensuring that the player encounters specific conditions during gameplay—for example, encountering \textit{Bat} and \textit{Spider} as required objectives.
The four story inputs used in this work are summarized in Table \ref{tab:scenario_examples}. 
We designed these scenarios to evaluate two complementary aspects: (1) the ability of the LLM to handle multiple conditional objectives, as in Scenarios 1 and 2, and (2) the capacity to reason about spatial constraints, as in Scenarios 3 and 4.

\begin{table}[h]
\centering
\caption{Examples of Scenario Instructions with Goal Flows}
\label{tab:scenario_examples}
\begin{tabular}{cp{7.5cm}}
\toprule
\textbf{\#} & \textbf{Scenario (Flow)} \\
\midrule
1 & \textit{The 'Player' needs to obtain the Key and escape through the 'Door'. To pick up the key, the player must encounter one of monsters: \textbf{BAT}.} \\
  & (Player $\rightarrow$ Bat $\rightarrow$ Key $\rightarrow$ Door) \\
\midrule
2 & \textit{The 'Player' needs to obtain the Key and escape through the 'Door'. To pick up the key, the player must encounter both of monsters: \textbf{BAT}, \textbf{SCORPION}.} \\
  & (Player $\rightarrow$ Bat \& Scorpion $\rightarrow$ Key $\rightarrow$ Door) \\
\midrule
3* & \textit{The player must pick up the \textbf{KEY} to unlock the door and discover the treasure inside.} \\
  & (Player $\rightarrow$ Key $\rightarrow$ Door $\rightarrow$ Treasure) \\
\midrule
4* & \textit{The \textbf{BAT} holds the key, and only by defeating it can the player open the door and reach the treasure.} \\
  & (Player $\rightarrow$ Bat $\rightarrow$  Door $\rightarrow$ Treasure) \\
\bottomrule \\
\multicolumn{2}{l}{\footnotesize *The player and treasure tiles are spatially separated by the door.}

\end{tabular}
\end{table}

We measure coherence-based accuracy by evaluating how well the player's experience along the path to the door aligns with the given instructional conditions.
To measure the game entities (keys and enemies) encountered by the player, we adopted a deterministic pathfinding algorithm for evaluation.
We evaluate the generated levels based on how well they align with specific gameplay scenarios, placing emphasis on ensuring a coherent player experience.

\section{Proposed Method}
\label{sec:method}

Our proposed framework \textit{PCGRLLM}, an improved reward generation framework for PCG, employs a three-step sequential approach: (1) refine the reward function through feedback, (2) align the reward function to the environment and train the agent, and (3) provide feedback to the reward function based on the generated content.
Fig. \ref{fig:architecture} illustrates the comprehensive architecture of the proposed framework.
While the prior study \cite{baek2024chatpcg} was the first to incorporate self-alignment into the reward generation task in content generation domain, this work extends the framework by incorporating feedback feature and a refinement process, forming an outer loop that enhances the overall system's adaptability and performance.
The following subsections detail the process of refining the reward function to align with the instruction inputs, with the corresponding pseudo-code provided in Algorithm \ref{alg:pcgrllm}.

\begin{algorithm}[!h]
\caption{PCGRLLM Reward Refinement Process}
\label{alg:pcgrllm}
\begin{algorithmic}[1]
\Require Task description $l$, LLM $\mathcal{M}$, fitness function $F$, environment $Env$, prompt inputs $p$

\Require Feedback count $N_{\text{feedback}}$, alignment count $N_{\text{align}}$

\State \textbf{Note:} \textcolor{blue}{blue color text} denotes variables used within the reasoning process.

\State $G \sim \text{InitializeGraph}(N_{\text{breadth}})$

\For{feedback step $y=1$ to $N_{\text{feedback}}$} 
    \State // \textbf{Step 1: Refinement (Section \ref{sec:reward_refinement})}
    \State $R, {Fb}, \textcolor{blue}{f} \sim G$ \Comment{Retrieve parent node.}
    \textcolor{blue}{
    \State $Aux = \{R^{\text{aux}}_{1:N_{\text{Best}}}, f^{\text{aux}}_{1:N_{\text{Best}}}\} \sim G$ \Comment{Retrieve auxiliary}
    }

    \State // \textbf{Step 2: Self-alignment (Section \ref{sec:self_alignment})}
    \State $R^{\prime} \sim \mathcal{M}(p_{\text{reward}}, l, R, {Fb}, \textcolor{blue}{f, Aux})$

    \For{alignment step $z=1$ to $N_{\text{align}}$} 
        \State $r^{\text{env}}_{1:N} \gets \text{Rollout}(\pi_{\text{random}},Env, R^{\prime})$
        \State $R^{\prime\prime} \gets \mathcal{N}(p_{\text{align}}, R^{\prime}, r^{\text{env}}_{1:M})$
        \State $R^{\prime} \gets R^{\prime\prime}$
    \EndFor

    \State $\pi_{\text{trained}} \gets \text{Train}(\pi_{\text{untrained}}, Env, R^{\prime})$
    \State $s_T \gets \text{Rollout}(\pi_{\text{trained}})$ \Comment{Generate content}
    \State $f \gets F(s_{T})$ \Comment{Evaluate fitness}

    \State // \textbf{Step 3: Feedback (Section \ref{sec:self_feedback})}
    \State ${Fb} \gets \mathcal{M}(p_{\text{feedback}}, l, R^{\prime}, s_T)$
    \State $G.\text{update}(R^{\prime}, s_T, f)$
\EndFor

\State \textbf{Output:} Refined reward function $R^{\prime}$
\end{algorithmic}
\end{algorithm}

\subsection{Reward Refinement}
\label{sec:reward_refinement}
The reward refinement process is an iterative procedure aimed at enhancing the reward function through feedback to better align it with the given instruction.
It takes as input a textual description of the game environment, the objective of the reward generation task, textual instructions, and accessible variables from the environment. For this study, two concise textual stories, as detailed in Section \ref{sec:story2reward}, were used as input for the textual instructions.
The process involves revising the parent function, defined as the previously generated reward function, and progresses iteratively, with each iteration denoted as \( y \). Depending on the presence of a prior iteration, the process operates in one of two phases: initializing the reward function or refining it based on feedback from previous iterations.

\textbf{Initial generation (\( y = 1 \))}
The initial generation phase begins with a template reward function, which is an empty function containing only parameter definitions. The parameters of the reward function are the previous and current level arrays, while the level size is intentionally excluded to prevent the LLM from generating hard-coded functions.
In the first iteration, the LLM generates conceptual ideas to outline the components of the reward function.
The prompt includes the progress of the feedback iteration, \textit{"\{current iter.\} of \{max iter.\}"}, to utilize the planning capabilities of LLMs.

\textbf{Continued refinement (\( y \geq 2 \))}
In subsequent iterations, the process retrieves a previously generated reward function from the reward function archive ($G$) to serve as the parent function for refinement. A pair consisting of the parent reward function (\( R \)) and its corresponding feedback ($Fb$) is sampled and provided to the LLM. The method for retrieving the parent reward function ($R$) varies depending on the prompt engineering method, as detailed in following section (Section \ref{sec:reasoning_pe}).

Once the parent function and feedback are identified, they are passed to the LLM ($\mathcal{M}$). Using a refinement prompt (\( p_\text{refine} \)), the LLM generates an improved reward function ($R'$) that incorporates the parent function and feedback.
The improved reward function addresses the shortcomings or problematic aspects of the previous reward function, making it better aligned with the textual instructions.

\subsection{Self-alignment to the Environment}
\label{sec:self_alignment}
The self-alignment process ensures that the generated reward function (\( R \)) produces the trainable reward signals in the environment, reflecting the intend of LLM.
By allowing the reward function to briefly interact with the training environment in a few-shot manner, this process refines the function to align with the designed insights.
Overly narrow reward ranges or excessively large reward values can hinder effective training DRL agent.
This iterative process enables the reward function to generate the intended content while providing meaningful feedback for each refinement.

Specifically, one episode is simulated using a random agent to collect distributed reward values directly from interactions with the environment, denoted as \( r^{\text{env}}_{1:M} \). The mean and variance of these reward values are calculated to verify whether the designed reward function outputs values within the intended range. Based on these evaluations, the LLM adjusts the reward function, producing an updated version (\( R'' \)) along with the actual reward values (\( r^{\text{env}}_{1:M} \)), the current reward function (\( R' \)), and the alignment prompt (\( p_{\text{align}} \)). This process, repeated \( N_{\text{align}} \) times, incrementally refines the reward function by modifying weights or formulas to consider the scale and sparsity of the reward signal.

\subsection{Feedback from Generated Contents}
\label{sec:self_feedback}

Feedback is an essential process for refining the reward function to reflect the actual output of the trained policy. The LLM reasons about the causal relationship between the reward function and the generated content to identify inconsistencies. It evaluates whether the generated content aligns with the instruct conditions by analyzing discrepancies, such as the positions or counts of important tiles within the level. Based on this analysis, the model formulates a plan to refine the reward function, guiding its next iteration toward better alignment with the intended design objectives and the policy's outputs.

First, the terminal states (\( s_T \)) are collected as generated content by inferring the trained policy (\( \pi_{\text{trained}} \)). Then, the LLM generates feedback by reasoning over the text-formatted levels or rendered images of the terminal states (\( s_T \)) along with the current reward function (\( R' \)) and a feedback prompt (\( p_{\text{feedback}} \)). To prevent hallucination and maintain precision, the number of feedback points is limited to one per iteration.

Once the feedback ($Fb$) is generated, the updated reward function and its associated feedback are added to the reward function archive (\( G \)) for use in the next iteration. This iterative process ensures that the reward function progressively aligns with both the intended objectives and the actual outputs of the trained policy, enabling the generation of increasingly optimal content.

\begin{figure*}[!t]
    \centering
    \begin{tabular}{cccccc} 
        \multicolumn{6}{l}{\small{Instruct: \textit{"The player needs to obtain a key and escape through the door. The player encounters \textbf{bat} and \textbf{spider} monsters."}}} \\
        \includegraphics[width=0.14\textwidth]{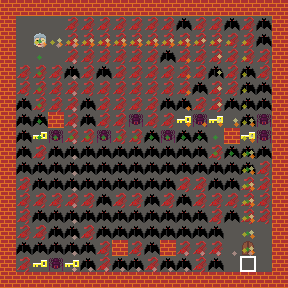} & 
        \includegraphics[width=0.14\textwidth]{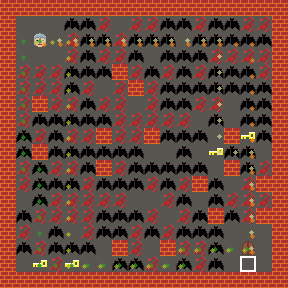} &
        \includegraphics[width=0.14\textwidth]{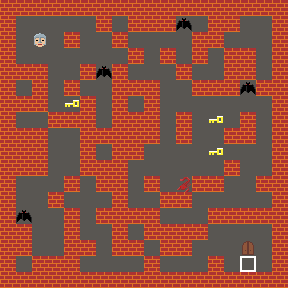} & 
        \includegraphics[width=0.14\textwidth]{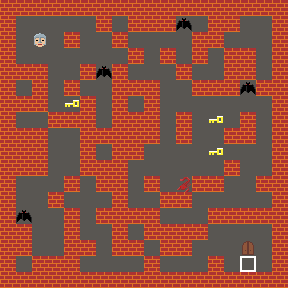} &
        \includegraphics[width=0.14\textwidth]{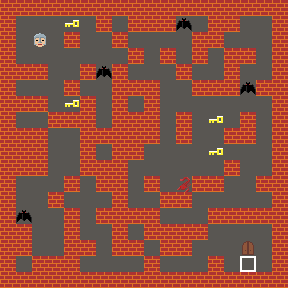} & 
        \includegraphics[width=0.14\textwidth]{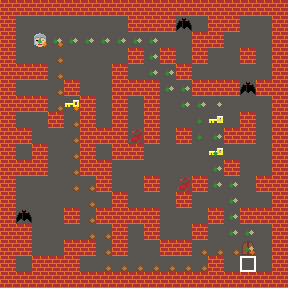} \\
        $*y=1$ & $y=2$ & $y=3$ & $*y=4$ & $y=5$ & $*y=6$ \\
        \multicolumn{6}{l}{\small{Instruct: \textit{"The player needs to obtain a key and escape through the door. The player encounters \textbf{bat} and \textbf{spider} monsters."}}} \\
        \includegraphics[width=0.14\textwidth]{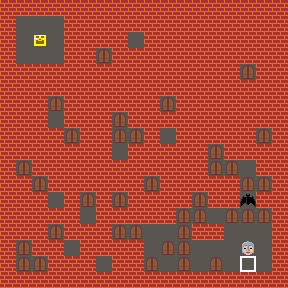} & 
        \includegraphics[width=0.14\textwidth]{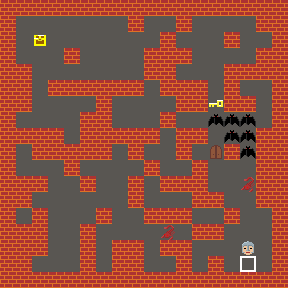} &
        \includegraphics[width=0.14\textwidth]{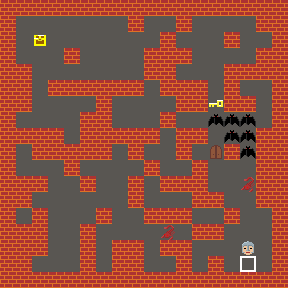} & 
        \includegraphics[width=0.14\textwidth]{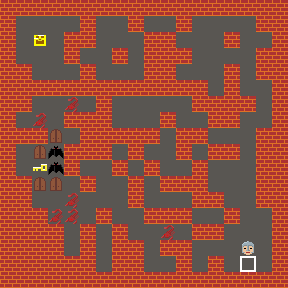} &
        \includegraphics[width=0.14\textwidth]{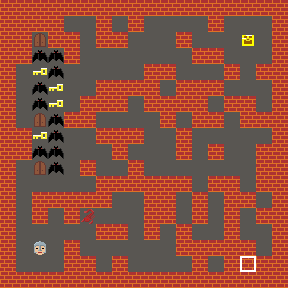} & 
        \includegraphics[width=0.14\textwidth]{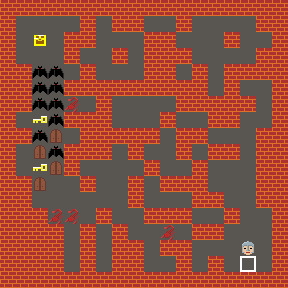} \\
        $y=1$ & $y=2$ & $y=3$ & $y=4$ & $*y=5$ & $*y=6$
    \end{tabular}
    \caption{The generated level images are from the iterative reward generation process based on the given instructions. Each map corresponds to an iteration ($y$), which represents the number of times the reward has been generated and revised by LLMs, and is produced by an agent trained using these LLM-generated reward functions.}
    \label{fig:iteration_examples}
\end{figure*}

\subsection{Reasoning-based Prompt Engineering for Reward Improving}
\label{sec:reasoning_pe}

We employed various state-of-the-art PE techniques to iteratively refine the reward function in a step-by-step manner. The process of generating reward functions is inherently a trial-and-error approach, which presents a significant challenge as it does not guarantee consistent improvements. 
Such an algorithm must effectively identify and address errors while maintaining the flexibility to explore alternative solutions, ensuring convergence toward improved reward designs.
To address this challenge, we adopted backtracking-enabled and branching PE methods, such as ToT and GoT, to expand the exploration of the reward space.
Fig. \ref{fig:reasoning_pe} illustrates the detailed procedure of thought node expansion and the utilization of auxiliary information in GoT.

\textbf{Thought node}
A feedback iteration unit is represented by a single thought node in the reasoning process, comprising three steps: reward refinement, self-alignment, and feedback. The expansion method varies depending on the PE approach. CoT employs a simple chain structure with a maximum breadth (\( N_{\text{breadth}} \)) of 1, while ToT and GoT utilize tree and graph structures, respectively, with a maximum breadth. For parent node selection, CoT expands from the latest unique node, whereas ToT and GoT expand from the node with the highest fitness score, provided its number of child nodes remains within the maximum breadth.

\textbf{Reward evaluation}
The fitness score, essential for determining the parent node in ToT and GoT, is measured within the range \([0, 1]\) and evaluates how well the generated content satisfies the instruction input. It can be determined using heuristics or self-evaluation with LLMs. In this study, we use a heuristic approach based on accuracy to mitigate the influence of subjective LLM evaluations. Specifically, we calculate the average accuracy of 30 instruct-level pairs.

\textbf{Auxiliary information}
GoT utilizes auxiliary inputs, incorporating the parent thought node along with the top-2 reward nodes and their corresponding fitness scores. This approach is motivated by the need to improve sample efficiency, as generating and training reward functions incur significant computational costs. By comparing the fitness value ($f$), the LLM can identify reward functions that best satisfy the instruct and discern their advantageous elements to combine them, facilitating more informed reasoning and improved outcomes.

Including prior work \cite{baek2024chatpcg}, most studies have predominantly adopted conventional prompt engineering strategies such as CoT, which emphasize stepwise reasoning but inherently constrain the exploration of the reward function space. In contrast, our approach leverages a broader set of state-of-the-art PE techniques, enabling more efficient searching on reward space.

\begin{table*}
\centering
\caption{
\textbf{Comparative performance across scenarios, reasoning methods, and module ablations.}
Results are reported as mean$\pm$standard deviation, with bold indicating the best value in each row.
The ``Mean'' scenario denotes the scenario-wise normalized mean score.
\textit{Self-alignment (SA)} and \textit{Feedback (FB)} correspond to ablation settings in which only the SA module or only the FB module of PCGRLLM is applied.
$p(\cdot)$ denotes the $t$-test $p$-value, with $p(\mathrm{ZS})$ and $p(\mathrm{SA})$ indicating comparisons of PCGRLLM against the zero-shot and SA settings, respectively.
}
\label{tab:performance_pcgrllm}
\begin{tabular}{p{1.5cm}p{1.5cm}p{2.6cm}p{2.6cm}p{2.6cm}p{2.6cm}cc}
\toprule
Scenario & Reasoning & Zero-shot (ZS) & +Self-alignment (SA) & +Feedback (FB) & PCGRLLM & $p$(ZS) & $p$(SA) \\
\midrule
 & CoT \cite{wei2022chain} & \cellcolor{gray!6}\rule{0pt}{2.2ex}\color{black} 0.064$\pm${\scriptsize 0.098} & \cellcolor{gray!5}\rule{0pt}{2.2ex}\color{black} 0.013$\pm${\scriptsize 0.012} & \cellcolor{gray!5}\rule{0pt}{2.2ex}\color{black} 0.011$\pm${\scriptsize 0.019} & \textbf{\cellcolor{gray!7}\rule{0pt}{2.2ex}\color{black} 0.093$\pm${\scriptsize 0.184}} & \rule{0pt}{2.2ex}0.767 & \rule{0pt}{2.2ex}0.387 \\
1 & ToT \cite{yao2024tree} & \cellcolor{gray!5}\rule{0pt}{2.2ex}\color{black} 0.004$\pm${\scriptsize 0.010} & \cellcolor{gray!6}\rule{0pt}{2.2ex}\color{black} 0.056$\pm${\scriptsize 0.088} & \textbf{\cellcolor{gray!13}\rule{0pt}{2.2ex}\color{black} 0.309$\pm${\scriptsize 0.379}} & \cellcolor{gray!9}\rule{0pt}{2.2ex}\color{black} 0.144$\pm${\scriptsize 0.123} & \rule{0pt}{2.2ex}0.063 & \rule{0pt}{2.2ex}0.229 \\
 & GoT \cite{besta2024graph} & \cellcolor{gray!5}\rule{0pt}{2.2ex}\color{black} 0.007$\pm${\scriptsize 0.006} & \cellcolor{gray!5}\rule{0pt}{2.2ex}\color{black} 0.002$\pm${\scriptsize 0.005} & \cellcolor{gray!5}\rule{0pt}{2.2ex}\color{black} 0.020$\pm${\scriptsize 0.005} & \textbf{\cellcolor{gray!7}\rule{0pt}{2.2ex}\color{black} 0.071$\pm${\scriptsize 0.130}} & \rule{0pt}{2.2ex}0.331 & \rule{0pt}{2.2ex}0.302 \\ 
\hdashline
 & Human & \rule{0pt}{2.2ex} 0.053$\pm${\scriptsize 0.054} & &  & \rule{0pt}{2.2ex} 1.000$\pm${\scriptsize 0.010} & &  \\ 
\midrule
 & CoT \cite{wei2022chain}& \cellcolor{gray!5}\rule{0pt}{2.2ex}\color{black} 0.002$\pm${\scriptsize 0.005} & \cellcolor{gray!5}\rule{0pt}{2.2ex}\color{black} 0.007$\pm${\scriptsize 0.010} & \textbf{\cellcolor{gray!13}\rule{0pt}{2.2ex}\color{black} 0.300$\pm${\scriptsize 0.322}} & \cellcolor{gray!9}\rule{0pt}{2.2ex}\color{black} 0.140$\pm${\scriptsize 0.313} & \rule{0pt}{2.2ex}0.381 & \rule{0pt}{2.2ex}0.395 \\
2 & ToT \cite{yao2024tree} & \cellcolor{gray!5}\rule{0pt}{2.2ex}\color{black} 0.000$\pm${\scriptsize 0.000} & \cellcolor{gray!5}\rule{0pt}{2.2ex}\color{black} 0.000$\pm${\scriptsize 0.000} & \textbf{\cellcolor{gray!12}\rule{0pt}{2.2ex}\color{black} 0.251$\pm${\scriptsize 0.303}} & \cellcolor{gray!7}\rule{0pt}{2.2ex}\color{black} 0.073$\pm${\scriptsize 0.110} & \rule{0pt}{2.2ex}0.211 & \rule{0pt}{2.2ex}0.211 \\
 & GoT \cite{besta2024graph} & \cellcolor{gray!5}\rule{0pt}{2.2ex}\color{black} 0.004$\pm${\scriptsize 0.010} & \cellcolor{gray!5}\rule{0pt}{2.2ex}\color{black} 0.002$\pm${\scriptsize 0.005} & \cellcolor{gray!9}\rule{0pt}{2.2ex}\color{black} 0.173$\pm${\scriptsize 0.291} & \textbf{\cellcolor{gray!12}\rule{0pt}{2.2ex}\color{black} 0.249$\pm${\scriptsize 0.281}} & \rule{0pt}{2.2ex}0.123 & \rule{0pt}{2.2ex}0.121 \\
\hdashline 
 & Human & \rule{0pt}{2.2ex} 0.371$\pm${\scriptsize 0.285} & &  & \rule{0pt}{2.2ex} 0.369$\pm${\scriptsize 0.055} & &  \\ 
 
\midrule
 & CoT \cite{wei2022chain} & \cellcolor{gray!5}\rule{0pt}{2.2ex}\color{black} 0.000$\pm${\scriptsize 0.000} & \cellcolor{gray!17}\rule{0pt}{2.2ex}\color{black} 0.420$\pm${\scriptsize 0.417} & \cellcolor{gray!5}\rule{0pt}{2.2ex}\color{black} 0.027$\pm${\scriptsize 0.028} & \textbf{\cellcolor{gray!27}\rule{0pt}{2.2ex}\color{black} 0.793$\pm${\scriptsize 0.361}}$^{\dagger}$ & \rule{0pt}{2.2ex}0.008 & \rule{0pt}{2.2ex}0.169 \\
3 & ToT \cite{yao2024tree} & \cellcolor{gray!5}\rule{0pt}{2.2ex}\color{black} 0.000$\pm${\scriptsize 0.000} & \cellcolor{gray!10}\rule{0pt}{2.2ex}\color{black} 0.187$\pm${\scriptsize 0.318} & \cellcolor{gray!5}\rule{0pt}{2.2ex}\color{black} 0.033$\pm${\scriptsize 0.075} & \textbf{\cellcolor{gray!16}\rule{0pt}{2.2ex}\color{black} 0.413$\pm${\scriptsize 0.536}} & \rule{0pt}{2.2ex}0.160 & \rule{0pt}{2.2ex}0.445 \\
 & GoT \cite{besta2024graph} & \cellcolor{gray!5}\rule{0pt}{2.2ex}\color{black} 0.013$\pm${\scriptsize 0.018} & \cellcolor{gray!5}\rule{0pt}{2.2ex}\color{black} 0.000$\pm${\scriptsize 0.000} & \cellcolor{gray!15}\rule{0pt}{2.2ex}\color{black} 0.360$\pm${\scriptsize 0.498} & \textbf{\cellcolor{gray!29}\rule{0pt}{2.2ex}\color{black} 0.873$\pm${\scriptsize 0.248}}$^{\dagger}$$^{*}$ & \rule{0pt}{2.2ex}0.001 & \rule{0pt}{2.2ex}0.001 \\  \hdashline
 & Human & \rule{0pt}{2.2ex} 0.000$\pm${\scriptsize 0.000} & &  & \rule{0pt}{2.2ex} 0.007$\pm${\scriptsize 0.015} & &  \\ 

\midrule
 & CoT \cite{wei2022chain} & \cellcolor{gray!5}\rule{0pt}{2.2ex}\color{black} 0.000$\pm${\scriptsize 0.000} & \textbf{\cellcolor{gray!7}\rule{0pt}{2.2ex}\color{black} 0.080$\pm${\scriptsize 0.145}} & \cellcolor{gray!5}\rule{0pt}{2.2ex}\color{black} 0.033$\pm${\scriptsize 0.047} & \cellcolor{gray!6}\rule{0pt}{2.2ex}\color{black} 0.067$\pm${\scriptsize 0.062} & \rule{0pt}{2.2ex}0.075 & \rule{0pt}{2.2ex}0.857 \\
4 & ToT \cite{yao2024tree} & \cellcolor{gray!5}\rule{0pt}{2.2ex}\color{black} 0.000$\pm${\scriptsize 0.000} & \cellcolor{gray!12}\rule{0pt}{2.2ex}\color{black} 0.253$\pm${\scriptsize 0.384} & \cellcolor{gray!5}\rule{0pt}{2.2ex}\color{black} 0.000$\pm${\scriptsize 0.000} & \textbf{\cellcolor{gray!21}\rule{0pt}{2.2ex}\color{black} 0.573$\pm${\scriptsize 0.438}}$^{\dagger}$ & \rule{0pt}{2.2ex}0.043 & \rule{0pt}{2.2ex}0.255 \\
 & GoT \cite{besta2024graph} & \cellcolor{gray!8}\rule{0pt}{2.2ex}\color{black} 0.107$\pm${\scriptsize 0.174} & \cellcolor{gray!5}\rule{0pt}{2.2ex}\color{black} 0.027$\pm${\scriptsize 0.060} & \cellcolor{gray!5}\rule{0pt}{2.2ex}\color{black} 0.000$\pm${\scriptsize 0.000} & \textbf{\cellcolor{gray!18}\rule{0pt}{2.2ex}\color{black} 0.467$\pm${\scriptsize 0.359}} & \rule{0pt}{2.2ex}0.092 & \rule{0pt}{2.2ex}0.051 \\
 \hdashline
 & Human & \rule{0pt}{2.2ex} 0.000$\pm${\scriptsize 0.000} & &  & \rule{0pt}{2.2ex} 0.033$\pm${\scriptsize 0.075} & &  \\ 
\midrule
 & CoT \cite{wei2022chain} & \cellcolor{gray!5}\rule{0pt}{2.2ex}\color{black} 0.010$\pm${\scriptsize 0.020} & \cellcolor{gray!9}\rule{0pt}{2.2ex}\color{black} 0.162$\pm${\scriptsize 0.207} & \cellcolor{gray!7}\rule{0pt}{2.2ex}\color{black} 0.099$\pm${\scriptsize 0.120} & \textbf{\cellcolor{gray!13}\rule{0pt}{2.2ex}\color{black} 0.281$\pm${\scriptsize 0.306}}$^{\dagger}$ & \rule{0pt}{2.2ex}0.029 & \rule{0pt}{2.2ex}0.348 \\
Mean & ToT \cite{yao2024tree} & \cellcolor{gray!5}\rule{0pt}{2.2ex}\color{black} 0.001$\pm${\scriptsize 0.003} & \cellcolor{gray!9}\rule{0pt}{2.2ex}\color{black} 0.157$\pm${\scriptsize 0.254} & \cellcolor{gray!9}\rule{0pt}{2.2ex}\color{black} 0.162$\pm${\scriptsize 0.189} & \textbf{\cellcolor{gray!15}\rule{0pt}{2.2ex}\color{black} 0.353$\pm${\scriptsize 0.310}}$^{\dagger}$ & \rule{0pt}{2.2ex}0.006 & \rule{0pt}{2.2ex}0.142 \\
 & GoT \cite{besta2024graph} & \cellcolor{gray!6}\rule{0pt}{2.2ex}\color{black} 0.053$\pm${\scriptsize 0.131} & \cellcolor{gray!5}\rule{0pt}{2.2ex}\color{black} 0.009$\pm${\scriptsize 0.022} & \cellcolor{gray!8}\rule{0pt}{2.2ex}\color{black} 0.135$\pm${\scriptsize 0.200} & \textbf{\cellcolor{gray!17}\rule{0pt}{2.2ex}\color{black} 0.420$\pm${\scriptsize 0.328}}$^{\dagger}$$^{*}$ & \rule{0pt}{2.2ex}0.010 & \rule{0pt}{2.2ex}0.006 \\ \hdashline
 & Human & \rule{0pt}{2.2ex} 0.110$\pm${\scriptsize 0.207} & &  & \rule{0pt}{2.2ex} 0.422$\pm${\scriptsize 0.531} & &  \\ 
\bottomrule \\
\multicolumn{8}{r}{\footnotesize Significance marks indicate $p<0.05$: $^{*}$ for +Self-alignment vs PCGRLLM, $^{\dagger}$ for Zero-shot vs PCGRLLM.}
\end{tabular}
\end{table*}

\section{Experiment}
To evaluate the PCGRLLM framework, we designed experiments that assess each module individually and formulated hypotheses to examine the limitations of LLMs.  
Specifically, we set the following research questions:

\begin{itemize}
    \item \textit{RQ1. Is the feedback mechanism effective in evolving the reward function?}
    \item \textit{RQ2. Does the quality of feedback influence the improvement of reward functions?}
    \item \textit{RQ3. Can the reasoning-based prompt engineering method improve the efficiency of exploration in reward design?}
    \item \textit{RQ4. Can self-evaluated fitness serve as a reliable criterion for guiding the search direction?}
\end{itemize}

\subsection{Experimental Setting}

\textbf{Reward generation} The number of self-feedback iterations was set to \( N_{\text{feedback}} = 6 \), and the self-alignment iterations were set to \( N_{\text{align}} = 5 \). For generating the LLM-based reward, OpenAI’s \texttt{gpt-4o-2024-08-06} \cite{hurst2024gpt} served as the primary backend language model. The breadth for the ToT and GoT methods was set to \( N_{\text{breadth}} = 2 \), balancing the depth of the thought nodes with exploration. The fitness function utilized an accuracy-based evaluation metric, calculated as the average score over 30 inferenced levels. To minimize variability in LLM responses, the temperature parameter, which governs stochasticity, was set to 0.
The prompts used in the experiments are noted in supplementary.

To validate the RQs, we conducted a series of controlled experiments. For RQ1, we performed an ablation of the feedback module to assess its contribution to reward evolution. For RQ2, we compared normal and predefined feedback to examine how feedback quality influences refinement. To evaluate RQ3, we tested three prompt engineering strategies—CoT, ToT, and GoT—focusing on their impact on exploration efficiency. Finally, for RQ4, we contrasted heuristic evaluation with LLM-based self-evaluated fitness to determine its reliability as a criterion for guiding search.

\textbf{Environment} The \textit{Dungeon} problem features seven game tiles and uses the \textit{Narrow} representation for the environment setting \cite{khalifa2020pcgrl}. The level size was set to \( 16 \times 16 \), and the agent was configured to scan the entire level three times per episode. The discrete action space consisted of five actions to modify the level, targeting five modifiable tiles: empty, wall, and three enemy types.
At the start of each episode, the player and one door were randomly placed along one of the four edges of the level, and the nearby \( 3 \times 3 \) tiles were masked as unmodifiable.

\textbf{RL training} The DRL models were trained using proximal policy optimization (PPO) \cite{schulman2017proximal} for 50 million timesteps using \textit{PureJaxRL} \cite{lu2022discovered} implementation and the hyperparameters detailed in Appendix \hyperref[sec:pcg_agent_parameters]{C}. 
To account for the inherent stochasticity of both the LLM and the DRL models, results were reported by aggregating across 5 random seeds.
All experimental results are reported as the mean and standard deviation across five independent runs; statistical significance is evaluated using a \textit{t}-test.
All experiments were conducted on RTX 8000 GPU machines.

\subsection{Evaluation Criteria}

\begin{figure}[!h]
    \centering

\begin{tabular}{cc}
    \includegraphics[width=0.45\linewidth]{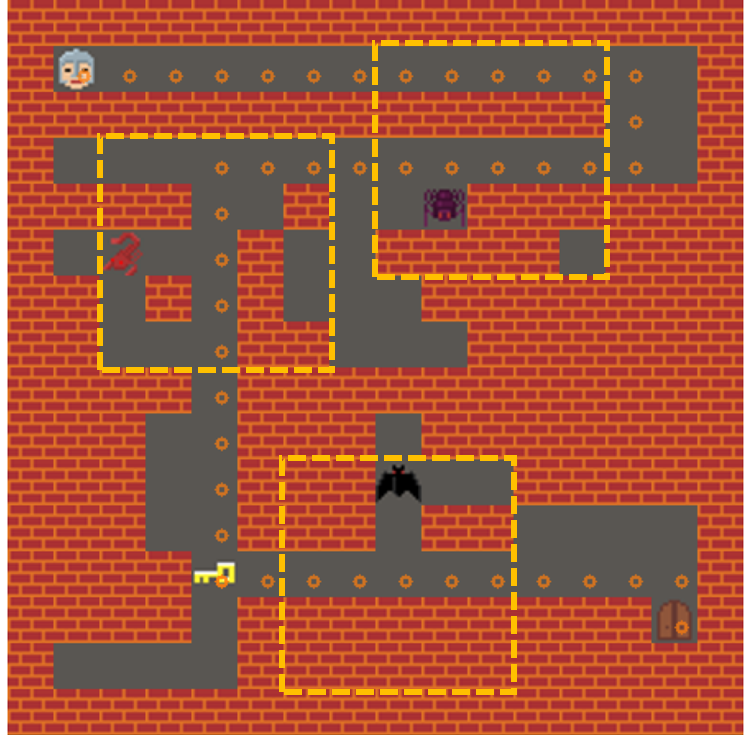} & 
    \includegraphics[width=0.45\linewidth]{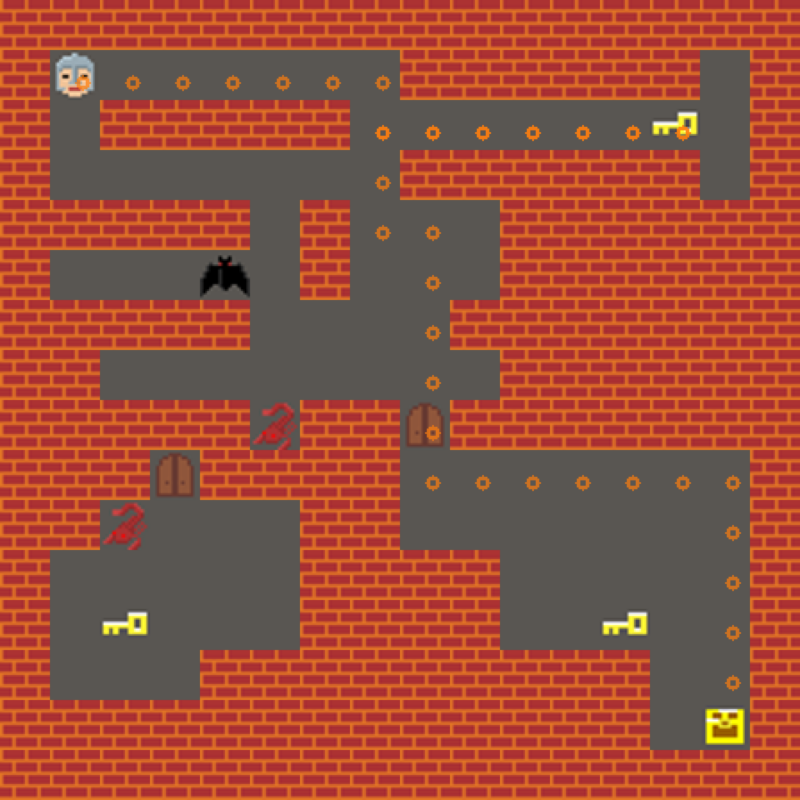} \\
    (a) Enemy encounter & (b) Key-door interaction
\end{tabular}

\caption{(a) The player encounters enemies within the yellow dotted box. 
(b) The player picks up a key to open the door.}
\label{fig:senario_method}
\end{figure}
\textbf{Enemy encounter scenario} measures whether the player encounters the specified key enemy tiles during gameplay, following the given scenario. 
We apply this metric in the context of scenario number 1 and 2.
We aim to evaluate how well the generated level satisfies the required enemy tiles specified in the instruction.
Let $E$ denote the set of enemy types (\textit{Bat}, \textit{Scorpion}, \textit{Spider}).
For each $e \in E$, the instruction defines whether it is required ($g_e = 1$) or not ($g_e = 0$).
During gameplay, the enemy encounter ($p_e$) is determined by whether the player actually encounters enemy $e$ along a viable path from the key to the door (identified using the flood-fill algorithm).
Enemies located within a 2-tile radius of the path are considered encountered.
To quantify instruction satisfaction, we compute:

\begin{equation}
\label{eq:satisfaction}
\text{Performance} = \frac{|\,\{e \in E \mid g_e = 1 \wedge p_e = 1\}\,|}{|\,\{e \in E \mid g_e = 1\}\,|}
\end{equation}

For example, if the instruction states, ``\textit{... the player encounters bat monsters,}'' and the generated level is shown in Fig. \ref{fig:senario_method}b, the ground truth is $[1, 0, 0]$, where only the \textit{Bat} is a positive label. 
If the prediction is $[1, 1, 0]$, where both \textit{Bat} and \textit{Scorpion} are marked as positive, the accuracy is calculated as $0.67$. 
The accuracy is measure on each instruction-level pair and the averaged with five runs.
The detailed evaluation algorithm is described in Appendix \hyperref[sec:acc_eval_method]{B}.

\textbf{Key-door interaction scenario} evaluates whether the generated level is playable under the key-door dependency, while also assessing the spatial separation between the player and the treasure. 
We apply this metric in the context of scenario number 3 and 4.
The scenario requires that the player and the treasure are separated by at least one locked door, and progression is only possible by collecting a reachable key. 
A level is considered playable if the player can reach a key from the starting position and subsequently unlock the corresponding door to access the treasure chest. 
If either the key is unreachable or the door does not lie on the path to the treasure, the level is considered unplayable.

To determine playability, we apply the flood-fill pathfinding algorithm to verify the existence of a valid route that respects the key-door spatial dependency. 
Fig.~\ref{fig:senario_method}a illustrates how the algorithm identifies viable paths from the player to the treasure, ensuring that the key is collected and the door is opened along the way. 
The evaluation thus focuses on binary playability (playable or unplayable) rather than accuracy scores, and results are averaged over five independent runs.

\subsection{Experimental Result}
\subsubsection{Effectiveness of Feedback (RQ1)}
\label{sec:effective_feedback}
Table \ref{tab:performance_pcgrllm} presents the ablation results by separating the self-alignment (SA) mechanism and the feedback module, where the SA condition corresponds to the prior method \textit{ChatPCG}~\cite{baek2024chatpcg}.
Zero-shot (ZS) setting serves as a baseline that relies solely on one-shot generation without any refinement mechanisms.
Statistical significance of these improvements was assessed using the Welch $t$-test \cite{ruxton2006unequal}, and the corresponding $p$-values are reported.

Across all PE, the introduction of feedback consistently improved performance compared to both zero-shot and SA-only settings, demonstrating its importance in aligning generated reward functions with the intended task objectives.  
PCGRLLM achieved the highest score in most cases, and although the SA condition occasionally showed stronger results, the feedback mechanism generally led to further improvements across the three PE settings (CoT, ToT, and GoT), with the most pronounced gains observed in the GoT condition.

These improvements are statistically significant over zero-shot across all PE ($p(\text{ZS})<0.05$), with further significant gains over SA in the GoT setting ($p(\text{SA})<0.01$).  
Overall, these results indicate that feedback not only mitigates the limitations of zero-shot generation but also provides clear benefits beyond the previously proposed self-alignment approach.  
In particular, \textbf{feedback enables the correction of shortcomings in agents trained with imperfect reward functions, thereby leading to more reliable and effective policy learning.}
The generalizability of the proposed framework was further verified using open-weight models, and the results are reported in Appendix~\hyperref[sec:open_weight_model_performance]{D}.

Compared with human-crafted reward design, a complementary pattern emerges: human-designed rewards perform better on tasks involving multiple conditional objectives (Scenarios 1–2), whereas PCGRLLM performs better on tasks requiring spatial reasoning under constraints (Scenarios 3–4). Appendix~\hyperlink{sec:human_crafted_reward}{H} further analyzes the factors underlying these differences, including scenario-wise reward sensitivity, trade-offs, and qualitative failure cases.
 
\subsubsection{Feedback Quality Ablation (RQ2)}
\label{sec:specificity_feedback}
To examine how feedback quality influences performance improvement, we compared conditions with and without feedback, as well as different levels of specificity in the feedback input.
This experiment benchmarks three feedback types—\textit{No}, \textit{Generic}, and \textit{Specific}—across multiple iterations.
Generic feedback provides only high-level guidance that is not directly related to the generated content, such as minor parameter tuning or superficial adjustments, while specific feedback serves as the default setting in this study.
Table~\ref{tab:iterwise_performance_got} reports the results for the GoT setting, showing that the specificity of feedback has a substantial effect on performance improvement over successive iterations.
Among the tested conditions, specific feedback achieved the highest performance by the sixth iteration.
By contrast, while \textit{Generic} feedback yielded some performance gains, neither reached the level of specific feedback, and the \textit{No} feedback condition in particular exhibited consistently weaker performance, underscoring the importance of feedback specificity.
Full experimental results are provided in Appendix~\hyperlink{sec:full_result_feedback}{E}.

\begin{table}
\centering
\caption{Iteration-wise performance for each feedback type. Bold values indicate the highest iteration per row.}
\label{tab:iterwise_performance_got}
\begin{tabular}{p{1.5cm} c c c c c c}
\toprule
Feedback & $y_{1}$ & $y_{2}$ & $y_{3}$ & $y_{4}$ & $y_{5}$ & $y_{6}$ \\
\midrule
No & \textbf{0.380} & 0.271 & 0.298 & 0.297 & 0.312 & 0.287 \\
Generic & 0.301 & 0.322 & 0.271 & 0.358 & 0.319 & \textbf{0.398} \\
Specific & 0.009 & 0.337 & 0.305 & 0.190 & 0.424 & \textbf{0.427} \\
\bottomrule
\end{tabular}
\end{table}

\begin{figure}[!h]
    \centering
    \includegraphics[width=\linewidth]{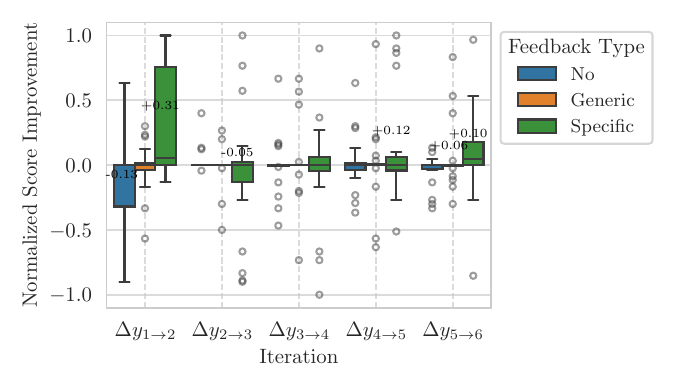}
    \caption{Performance change ($\Delta y$) across iterations for the three feedback types. Each bar represents the relative change ($\Delta y_{n \to n+1}$) between consecutive iterations.}
    \label{fig:specific_feedback}
\end{figure}

Fig.~\ref{fig:specific_feedback} illustrates the relative accuracy change ($\Delta y$) across iterations for each feedback type, further investigating how feedback influences performance by computing run-wise changes in accuracy across iterations.
The results indicate that the \textit{No} feedback condition generally leads to performance deline with little change across iterations, while the \textit{Generic} feedback condition shows only minimal variation without substantial improvement.
In contrast, \textit{Specific} feedback produces clear improvements in several iterations, especially in the early stages, underscoring its effectiveness in driving meaningful iterative refinement.
Overall, these findings demonstrate that \textbf{content-tailored, specific feedback plays a critical role in enhancing reward function quality and enabling sustained performance improvement}.

\subsubsection{Effectiveness of Reasoning Prompt Engineering (RQ3)}
To compare how PCGRLLM searches for reward functions, we evaluated three prompting strategies—CoT, ToT, and GoT—as baselines. 
Table \ref{tab:reasoning_prompt} reports the scenario-wise normalized mean scores across six iterations.
As shown in the table, performance does not monotonically improve with more iterations; therefore, overcoming score deline is essential for enabling efficient exploration.

\begin{table}[!h]
\centering
\caption{Iteration-wise performance between reasoning-based prompt engineering methods.}
\label{tab:reasoning_prompt}
\begin{tabular}{p{1.8cm}|rrrrrr}
\toprule
  \textbf{PE} & \multicolumn{6}{c}{Iteration ($y_{i}$)} \\
 & 1 & 2 & 3 & 4 & 5 & 6 \\
\midrule
CoT & 0.162 & 0.200 & \textbf{0.370} & 0.205 & 0.162 & 0.287 \\
ToT & 0.160 & 0.191 & 0.194 & 0.344 & 0.342 & \textbf{0.361} \\
GoT & 0.009 & 0.337 & 0.305 & 0.190 & 0.424 & \textbf{0.427} \\
\bottomrule
\end{tabular}
\end{table}
Although no statistically significant differences were found, the analysis revealed a descriptive trend. 
At the six iteration, GoT showed a tendency to outperform CoT ($p=0.114$) and ToT ($p=0.348$).
Although GoT shows a temporary drop at iteration 4 (0.190), its backtracking mechanism enables efficient exploration, ultimately achieving the highest performance (0.427). 
In contrast, CoT reaches its peak at iteration 3 (0.370) but fails to recover from subsequent performance degradation caused by misguided exploration, resulting in lower final scores than GoT. 
These results suggest that \textbf{advanced PE techniques such as backtracking and sample efficiency play a crucial role in enhancing the performance}

\subsubsection{Self-evaluation of Reward Fitness (RQ4)}

Fitness plays a crucial role in tree- and graph-based PE as it guides the search direction during reward optimization. 
However, most LLMs still suffer from potential issues such as hallucination, which has led prior studies \cite{yao2024tree,besta2024graph} to rely heavily on heuristic functions as evaluators. 
To overcome this limitation, it is necessary for LLMs to perform autonomous and objective evaluation of game content.
Concretely, for each input scenario, the LLM received a set of ten generated levels and generate a scenario-conditioned scalar score in $[0,1]$.
The results presented in Table \ref{tab:self_evaluation} highlight a clear contrast between the heuristic (oracle) and LLM-based evaluation methods.

\begin{table}[!h]
\caption{Performance comparison of two fitness functions, heuristic (oracle) and LLM-based self-evaluation.}
\label{tab:self_evaluation}
\begin{tabular}{p{1.4cm}|p{0.73cm}p{0.73cm}p{0.73cm}|p{0.73cm}p{0.73cm}p{0.73cm}}
\toprule
\textbf{Fitness ($F$)} & \multicolumn{3}{c|}{Heuristic (Oracle)} & \multicolumn{3}{c}{LLM (Self-evaluate)} \\
 & ZS & +FB & $\Delta{}$ & ZS & +FB & $\Delta{}$ \\
PE &  &  &  &  &  &  \\
\midrule
ToT & 0.001 & 0.163 & +0.162 & 0.261 & 0.050 & -0.211 \\
GoT & 0.053 & 0.142 & +0.090 & 0.125 & 0.069 & -0.056 \\
\bottomrule
\end{tabular}
\end{table}

The heuristic (oracle) evaluation consistently benefited from feedback, improving by +0.162 in ToT and +0.090 in GoT. 
In contrast, the LLM self-evaluation showed performance degradation, with a decline of -0.211 in ToT and -0.056 in GoT.
These results indicate that inaccurate evaluation scores from the LLM often guided the search toward suboptimal directions, thereby reducing the effectiveness of feedback. 
These findings indicate that \textbf{LLM-based self-evaluation, while currently limited in providing reliable guidance for efficient reward space exploration}, still holds promise as a direction for future improvement and investigation.

\section{Discussion}

\subsection{Auxiliary Information Analysis}
\label{sec:aux_info_analysis}

\begin{figure}[!h]
    \centering
    \includegraphics[width=0.8\linewidth]{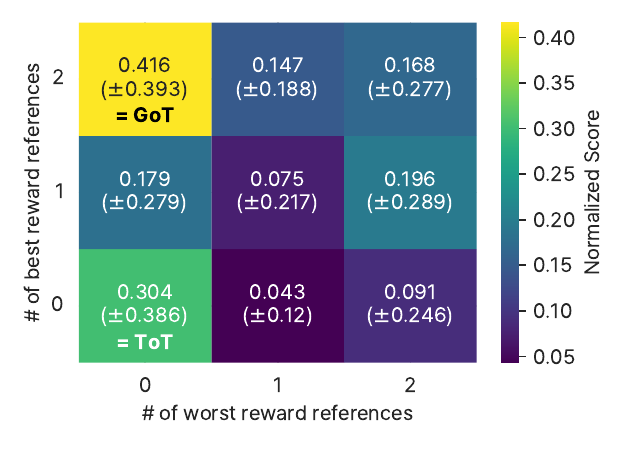}
    \caption{Normalized performance scores based on the number of auxiliary data in the GoT feedback process, categorized by best- (rows) and worst- (columns) fitness thoughts.}
    \label{fig:auxiliary_data_heatmap}
\end{figure}

Few-shot inputs strongly influence the generation quality of language models. In the GoT framework, incorporating the two best-performing nodes as reference data proved effective for shaping reward functions. To assess whether the number or quality of few-shot examples is more critical, we analyze how different input types affect generation outcomes (Fig. \ref{fig:auxiliary_data_heatmap}). Best nodes correspond to the top-2 highest-reward nodes, while worst nodes denote the lowest-reward ones. High-quality thoughts (high fitness values) are expected to expand the upper bound of performance, whereas low-quality thoughts (low fitness values) act as negative signals that help prevent performance collapse toward the lower bound.

Few-shot reference types influence generation outcomes in nuanced ways, offering insight into how models respond to varying input quality. The heatmap does not show statistically significant differences, but certain trends can be observed.
The existence of best nodes generally demonstrates a positive trend in performance improvement, although the magnitude of this effect is not always substantial. In contrast, the existence of worst nodes consistently results in performance degradation, thereby underscoring their detrimental influence.
Interestingly, even when scores are provided alongside the examples, the model fails to distinguish the flawed aspects of worst nodes and thus cannot raise the lower bound of performance. This implies that the critical factor is not the sheer number of few-shot examples but rather the preprocessing method or quality of feedback used to guide them.

\subsection{Integration with Diverse PCGRL Domains}
\label{sec:extened_pcgrl}
The proposed framework is not restricted to two-dimensional tile-based environments and can be extended to diverse PCGRL domains. In particular, it can be generalized to three-dimensional environments by defining rewards with 3D-specific criteria, such as spatial validity and structural consistency, as explored in a prior 3D PCGRL study \cite{jiang2022learning}. Similar reasoning-based prompting techniques may also support DRL-driven balancing tasks in multiplayer game settings, as suggested by prior work on LLM-assisted reward design \cite{baek2024chatpcg}. Moreover, when explicit symbolic metrics are difficult to design, vision-based models can provide feedback directly from rendered outputs, enabling reward signals derived from the visual properties of generated structures or objects. The approach is also compatible with simulation-driven evaluation, where rewards are computed from agent performance or gameplay outcomes.

\subsection{Extensions to Search-based Generation and Curriculum Learning}
\label{sec:extened_frameowork}

Reward functions (or objective functions) are central to both RL and search-based algorithms, as they define target behavior and shape the exploration and evaluation dynamics, and improvements in their design are expected to be beneficial in both domains \cite{summerville2018procedural}. 
The proposed framework can also be extended to search-based generation methods such as genetic algorithms or Monte Carlo tree search. In these algorithms, objective functions similarly govern the exploration dynamics of the search process, and iterative feedback can be used to refine these functions to guide the search toward semantically meaningful and higher-quality levels. 

Furthermore, the feedback mechanism introduced in PCGRLLM also suggests potential synergy with curriculum learning paradigms. In curriculum learning, tasks are organized with increasing levels of difficulty, allowing agents to progressively acquire skills. The feedback loop naturally supports this paradigm by identifying shortcomings of current reward functions and adjusting them to scaffold more achievable intermediate objectives. This could facilitate smoother training dynamics, reduce sparse reward issues, and enable gradual progression toward complex content generation goals.

\subsection{Reward Generation vs. End-to-end LLM Generation}
\label{sec:end2end}
In this work, large language models are used not as generators of content but as designers of reward functions. This formulation provides explicit and composable control over the learning process, as high-level preferences such as validity, diversity, novelty, safety, and style are mapped into interpretable reward terms with tunable weights. Such a design enables reliable enforcement of constraints, transparent trade-offs among competing objectives, reuse of reward functions across tasks and datasets, and systematic auditing of the factors that drive agent behavior. Within this paradigm, the LLM functions as a critic or author of reward functions, scoring, proposing edits, and reweighting terms rather than serving as the generator itself. This separation preserves low-latency inference, mitigates direct LLM biases during sampling, and enhances stability and reproducibility. Consequently, assigning generation to the policy while delegating reward refinement to the LLM yields finer controllability, improved generalization under distribution shift, and a clearer path to safety and compliance compared to end-to-end LLM-based generation.

\section{Conclusion and Future Work}
This study introduces an advanced reward generation architecture for game content generation.
PCGRLLM frames the pre-train and post-train reward refinement processes in terms of self-alignment and feedback, and its performance is evaluated across four distinct scenarios.
The results demonstrate that feedback-based reflection significantly improves reward functions, underscoring the crucial role of feedback quality. Furthermore, the experiments show that advanced PE methods enhance sample efficiency through mechanisms such as backtracking. The generality of the framework is validated using two popular foundation LLMs, highlighting its potential for integration with search-based baselines and applicability to broader game domains.

A key limitation of this study lies in the self-evaluation approach, where the LLM directly assessed the quality of generated content. 
The experiment revealed that such evaluation frequently misled the reward search process, as the scores produced by the LLM lacked sufficient objectivity and consistency.
To address this challenge, few-shot generation can provide comparative reference points for scoring, while self-consistency reasoning method could reduce the hallucination on evaluation. These approaches are expected to yield more objective assessment signals, enhance the robustness of the refinement process, and ultimately improve the effectiveness of reward space exploration.

\section{Acknowledgment}
This work was supported by the National Research Foundation of Korea (NRF) grant funded by the Korea government (MSIT) (RS-2025-16902996).
This work was supported by Institute of Information \& communications Technology Planning \& Evaluation (IITP) grant funded by the Korea government (MSIT) (No.2019-0-01842, Artificial Intelligence Graduate School Program (GIST)).
This work was supported by GIST-IREF from Gwangju Institute of Science and Technology (GIST).

\bibliographystyle{IEEEtran}
\bibliography{references}




\newpage

\section*{Appendix}
\label{sec:appendix}
\subsection{The 2D Level Generation Environment}

This study uses the \textit{PCGRL-Jax} \cite{earle2024scaling} environment, a GPU-accelerated implementation of the widely used two-dimensional level generation framework \cite{khalifa2020pcgrl,earle2021learning}.
The selected environment ensures a deterministic reward setting compared to the stochastic reward signal used in the previous study \cite{baek2024chatpcg}, so that the generated reward function is relatively accurately evaluated with a consistently trained policy.
Moreover, the environment provides 17x faster training time than CPU-based environment for multiple reward generate-and-evaluate iterations.

Each episode begins with a randomly initialized 16 $\times{}$ 16 matrix derived from a predefined tile set. The tile set consists of seven types: \textit{Empty} \includegraphics[height=0.8em]{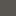}, \textit{Wall} \includegraphics[height=0.8em]{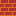}, \textit{Player} \includegraphics[height=0.8em]{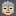}, \textit{Bat} \includegraphics[height=0.8em]{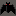}, \textit{Scorpion} \includegraphics[height=0.8em]{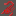}, \textit{Spider} \includegraphics[height=0.8em]{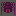}, \textit{Key} \includegraphics[height=0.8em]{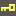}, \textit{Door} \includegraphics[height=0.8em]{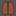}, and \textit{Treasure}\includegraphics[height=0.8em]{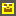}. Each tile type is represented numerically in the matrix to indicate its presence.
The agent can modify five types of tiles, except for two unchangeable \textit{Player} and \textit{Door} tiles, along with the $3 \times 3$ area of unchangeable tiles surrounding the tiles.
The two unchangeable tiles are randomly spawned on the opposite corners of the level in the initial state.
The observation space is defined as a 2D array representing the integer tile numbers, along with a channel the location of the tile to be modified.
The discrete action space includes five actions, each corresponding to the specific tile type that replaces the tile at the modification location.
The reward for the agent is determined by an LLM-generated reward function, implemented using JAX-compatible functions \cite{jax2018github}.

\subsection{Accuracy Evaluation Method}
\label{sec:acc_eval_method}

\definecolor{key_1}{rgb}{0.0, 0.0, 0.6}
\definecolor{key_2}{rgb}{0.5, 0.3, 0.0}
\definecolor{key_3}{rgb}{0.0, 0.3, 0.0}

The accuracy measurement involves two sequential steps: (1) identifying solution trajectories and (2) determining the types of encountered enemies. The encountered enemies are identified based on the solution trajectories, simulating player traversal along these paths. Fig. \ref{fig:senario_method}a illustrates the process of determining solution trajectories. There are three possible solutions corresponding to the three keys in the level. The trajectory for \textcolor{key_1}{\textit{Key 1} (blue)} passes through \textcolor{key_2}{\textit{Key 2} (brown)}, rendering it ineligible as an independent solution. The algorithm excludes duplicated paths and prioritizes the shortest paths, based on the assumption that players prefer the shortest traversal routes. In contrast, the trajectories leading to \textcolor{key_2}{\textit{Key 2} (brown)} and \textcolor{key_3}{\textit{Key 3} (green)} do not overlap with those of other keys, making them independent solutions.
As a result, the number of solutions is two, corresponding to \textcolor{key_2}{\textit{Key 2} (brown)} and \textcolor{key_3}{\textit{Key 3} (green)}.

\begin{algorithm}[!h]
\caption{Enemy Encounter Detection Logic}
\label{alg:distinct_path}
\textbf{Require:} Path finding algorithm $\mathcal{P}$, enemy tiles $E$ \\
\textbf{Require:} Player position $P$, door position $D$, key position list $K$, generated level $S_{T}$

\begin{algorithmic}[1]
\State $p \gets [\text{False}, \text{False}, \text{False}]$ \Comment{Initialize prediction array}
\For{key $k \in K$}
    \State $\tau_{S\to k} \gets \mathcal{P}(S, k), \quad \tau_{k\to D} \gets \mathcal{P}(k, D)$
    \If{$\tau_{S\to k} = \emptyset \ \textbf{or}\ \tau_{k\to D} = \emptyset \ \textbf{or}\ \sum_{x \in \tau_{S\to k}} [x \in K] \neq 1$}
        \State \textbf{continue} \Comment{Not connected or duplicated path}
    \EndIf
    \State $\tau_{P\to D} \gets \tau_{S\to k} + \tau_{k\to D}$ \Comment{Solution trajectory}
    \State $k_5 \gets \text{5x5 kernel to match enemy tiles}$
    \For{$(x, y) \in \tau_{P\to D}$}
        \State $\text{kernel\_tiles} \gets \{S_{T}[x+i, y+j] \mid (i, j) \in k_5\}$
        \For{$t \in \text{kernel\_tiles}$}
            \If{$t \in E$}
                \State $p[t] \gets \text{True}$ \Comment{Update prediction}
            \EndIf
        \EndFor
    \EndFor
\EndFor
\State \textbf{return} Encountered enemies $p$
\end{algorithmic}
\end{algorithm}

Algorithm \ref{alg:distinct_path} describes the process of detecting enemies encountered from the player to door positions. For each key $k \in K$, the algorithm calculates the path $\tau_{S \to k}$ from the start $S$ to the key and the path $\tau_{k \to D}$ from the key to the door $D$ using the flood-fill pathfinding algorithm.
Determine whether there is connectivity between the player, key, and door, and whether there is a single unique key on the path, defining this as the solution trajectory. Using the coordinates of the trajectory path, perform a convolution operation with a 5 by 5-sized kernel to detect the presence of enemies. Update the prediction ($p$) based on the types of detected enemies.

\begin{table*}
\centering
\caption{
\textbf{Comparative performance across scenarios, reasoning methods, and module ablations in the open-weight model.} 
Results are reported as mean$\pm$standard deviation, with bold indicating the best value in each row.
The ``Mean'' scenario denotes the scenario-wise normalized mean score.
\textit{Self-alignment (SA)} and \textit{Feedback (FB)} correspond to ablation settings in which only the SA module or only the FB module of PCGRLLM is applied.
$p(\cdot)$ denotes the $t$-test $p$-value, with $p(\mathrm{ZS})$ and $p(\mathrm{SA})$ indicating comparisons of PCGRLLM against the zero-shot and SA settings, respectively.
}
\label{tab:performance_llama}
\begin{tabular}{p{1.5cm}p{1.5cm}p{2.6cm}p{2.6cm}p{2.6cm}p{2.6cm}cc}
\toprule
Scenario & Reasoning & Zero-shot (ZS) & +Self-alignment (SA) & +Feedback (FB) & PCGRLLM &  $p$(ZS) & $p$(SA) \\
\midrule
 & CoT \cite{wei2022chain} & \cellcolor{gray!9}\rule{0pt}{2.2ex}\color{black} 0.147$\pm${\scriptsize 0.089} & \cellcolor{gray!13}\rule{0pt}{2.2ex}\color{black} 0.273$\pm${\scriptsize 0.084} & \cellcolor{gray!5}\rule{0pt}{2.2ex}\color{black} 0.027$\pm${\scriptsize 0.048} & \textbf{\cellcolor{gray!13}\rule{0pt}{2.2ex}\color{black} 0.293$\pm${\scriptsize 0.083}}$^{\dagger}$ & \rule{0pt}{2.2ex}0.028 & \rule{0pt}{2.2ex}0.716 \\
1 & ToT \cite{yao2024tree} & \cellcolor{gray!12}\rule{0pt}{2.2ex}\color{black} 0.267$\pm${\scriptsize 0.216} & \textbf{\cellcolor{gray!14}\rule{0pt}{2.2ex}\color{black} 0.333$\pm${\scriptsize 0.000}} & \cellcolor{gray!8}\rule{0pt}{2.2ex}\color{black} 0.109$\pm${\scriptsize 0.122} & \cellcolor{gray!11}\rule{0pt}{2.2ex}\color{black} 0.220$\pm${\scriptsize 0.159} & \rule{0pt}{2.2ex}0.708 & \rule{0pt}{2.2ex}0.187 \\
 & GoT \cite{besta2024graph} & \textbf{\cellcolor{gray!13}\rule{0pt}{2.2ex}\color{black} 0.284$\pm${\scriptsize 0.295}} & \cellcolor{gray!12}\rule{0pt}{2.2ex}\color{black} 0.262$\pm${\scriptsize 0.100} & \cellcolor{gray!7}\rule{0pt}{2.2ex}\color{black} 0.069$\pm${\scriptsize 0.065} & \cellcolor{gray!10}\rule{0pt}{2.2ex}\color{black} 0.198$\pm${\scriptsize 0.118} & \rule{0pt}{2.2ex}0.567 & \rule{0pt}{2.2ex}0.380 \\
\midrule
 & CoT \cite{wei2022chain} & \cellcolor{gray!18}\rule{0pt}{2.2ex}\color{black} 0.453$\pm${\scriptsize 0.261} & \cellcolor{gray!14}\rule{0pt}{2.2ex}\color{black} 0.322$\pm${\scriptsize 0.296} & \cellcolor{gray!18}\rule{0pt}{2.2ex}\color{black} 0.451$\pm${\scriptsize 0.349} & \textbf{\cellcolor{gray!24}\rule{0pt}{2.2ex}\color{black} 0.651$\pm${\scriptsize 0.397}} & \rule{0pt}{2.2ex}0.383 & \rule{0pt}{2.2ex}0.179 \\
2 & ToT \cite{yao2024tree} & \cellcolor{gray!14}\rule{0pt}{2.2ex}\color{black} 0.340$\pm${\scriptsize 0.307} & \cellcolor{gray!10}\rule{0pt}{2.2ex}\color{black} 0.191$\pm${\scriptsize 0.263} & \cellcolor{gray!15}\rule{0pt}{2.2ex}\color{black} 0.362$\pm${\scriptsize 0.378} & \textbf{\cellcolor{gray!19}\rule{0pt}{2.2ex}\color{black} 0.507$\pm${\scriptsize 0.257}} & \rule{0pt}{2.2ex}0.380 & \rule{0pt}{2.2ex}0.091 \\
 & GoT \cite{besta2024graph} & \cellcolor{gray!5}\rule{0pt}{2.2ex}\color{black} 0.013$\pm${\scriptsize 0.030} & \cellcolor{gray!15}\rule{0pt}{2.2ex}\color{black} 0.353$\pm${\scriptsize 0.293} & \cellcolor{gray!20}\rule{0pt}{2.2ex}\color{black} 0.544$\pm${\scriptsize 0.343} & \textbf{\cellcolor{gray!21}\rule{0pt}{2.2ex}\color{black} 0.569$\pm${\scriptsize 0.358}}$^{\dagger}$ & \rule{0pt}{2.2ex}0.025 & \rule{0pt}{2.2ex}0.329 \\
\midrule
 & CoT \cite{wei2022chain} & \cellcolor{gray!13}\rule{0pt}{2.2ex}\color{black} 0.293$\pm${\scriptsize 0.237} & \cellcolor{gray!13}\rule{0pt}{2.2ex}\color{black} 0.307$\pm${\scriptsize 0.245} & \cellcolor{gray!17}\rule{0pt}{2.2ex}\color{black} 0.420$\pm${\scriptsize 0.305} & \textbf{\cellcolor{gray!26}\rule{0pt}{2.2ex}\color{black} 0.733$\pm${\scriptsize 0.220}}$^{\dagger}$$^{*}$ & \rule{0pt}{2.2ex}0.016 & \rule{0pt}{2.2ex}0.020 \\
3 & ToT \cite{yao2024tree} & \cellcolor{gray!20}\rule{0pt}{2.2ex}\color{black} 0.527$\pm${\scriptsize 0.253} & \cellcolor{gray!24}\rule{0pt}{2.2ex}\color{black} 0.667$\pm${\scriptsize 0.217} & \cellcolor{gray!15}\rule{0pt}{2.2ex}\color{black} 0.373$\pm${\scriptsize 0.292} & \textbf{\cellcolor{gray!29}\rule{0pt}{2.2ex}\color{black} 0.853$\pm${\scriptsize 0.137}}$^{\dagger}$ & \rule{0pt}{2.2ex}0.043 & \rule{0pt}{2.2ex}0.150 \\
 & GoT \cite{besta2024graph} & \cellcolor{gray!17}\rule{0pt}{2.2ex}\color{black} 0.413$\pm${\scriptsize 0.221} & \cellcolor{gray!19}\rule{0pt}{2.2ex}\color{black} 0.507$\pm${\scriptsize 0.366} & \textbf{\cellcolor{gray!24}\rule{0pt}{2.2ex}\color{black} 0.680$\pm${\scriptsize 0.279}} & \cellcolor{gray!23}\rule{0pt}{2.2ex}\color{black} 0.620$\pm${\scriptsize 0.307} & \rule{0pt}{2.2ex}0.260 & \rule{0pt}{2.2ex}0.611 \\
\midrule
 & CoT \cite{wei2022chain} & \cellcolor{gray!5}\rule{0pt}{2.2ex}\color{black} 0.000$\pm${\scriptsize 0.000} & \cellcolor{gray!5}\rule{0pt}{2.2ex}\color{black} 0.027$\pm${\scriptsize 0.028} & \cellcolor{gray!16}\rule{0pt}{2.2ex}\color{black} 0.387$\pm${\scriptsize 0.455} & \textbf{\cellcolor{gray!26}\rule{0pt}{2.2ex}\color{black} 0.727$\pm${\scriptsize 0.386}}$^{\dagger}$$^{*}$ & \rule{0pt}{2.2ex}0.014 & \rule{0pt}{2.2ex}0.015 \\
4 & ToT \cite{yao2024tree} & \cellcolor{gray!5}\rule{0pt}{2.2ex}\color{black} 0.000$\pm${\scriptsize 0.000} & \cellcolor{gray!5}\rule{0pt}{2.2ex}\color{black} 0.027$\pm${\scriptsize 0.060} & \textbf{\cellcolor{gray!11}\rule{0pt}{2.2ex}\color{black} 0.207$\pm${\scriptsize 0.444}} & \cellcolor{gray!10}\rule{0pt}{2.2ex}\color{black} 0.193$\pm${\scriptsize 0.251} & \rule{0pt}{2.2ex}0.160 & \rule{0pt}{2.2ex}0.215 \\
 & GoT \cite{besta2024graph} & \cellcolor{gray!5}\rule{0pt}{2.2ex}\color{black} 0.007$\pm${\scriptsize 0.015} & \cellcolor{gray!5}\rule{0pt}{2.2ex}\color{black} 0.007$\pm${\scriptsize 0.015} & \cellcolor{gray!11}\rule{0pt}{2.2ex}\color{black} 0.233$\pm${\scriptsize 0.394} & \textbf{\cellcolor{gray!16}\rule{0pt}{2.2ex}\color{black} 0.407$\pm${\scriptsize 0.409}} & \rule{0pt}{2.2ex}0.094 & \rule{0pt}{2.2ex}0.094 \\
\midrule
 & CoT \cite{wei2022chain} & \cellcolor{gray!11}\rule{0pt}{2.2ex}\color{black} 0.226$\pm${\scriptsize 0.145} & \cellcolor{gray!12}\rule{0pt}{2.2ex}\color{black} 0.242$\pm${\scriptsize 0.105} & \cellcolor{gray!15}\rule{0pt}{2.2ex}\color{black} 0.352$\pm${\scriptsize 0.157} & \textbf{\cellcolor{gray!22}\rule{0pt}{2.2ex}\color{black} 0.598$\pm${\scriptsize 0.282}}$^{\dagger}$$^{*}$ & \rule{0pt}{2.2ex}0.007 & \rule{0pt}{2.2ex}0.009 \\
Mean & ToT \cite{yao2024tree} & \cellcolor{gray!13}\rule{0pt}{2.2ex}\color{black} 0.278$\pm${\scriptsize 0.201} & \cellcolor{gray!13}\rule{0pt}{2.2ex}\color{black} 0.303$\pm${\scriptsize 0.247} & \cellcolor{gray!11}\rule{0pt}{2.2ex}\color{black} 0.222$\pm${\scriptsize 0.198} & \textbf{\cellcolor{gray!18}\rule{0pt}{2.2ex}\color{black} 0.466$\pm${\scriptsize 0.281}} & \rule{0pt}{2.2ex}0.109 & \rule{0pt}{2.2ex}0.185 \\
 & GoT \cite{besta2024graph} & \cellcolor{gray!10}\rule{0pt}{2.2ex}\color{black} 0.200$\pm${\scriptsize 0.202} & \cellcolor{gray!13}\rule{0pt}{2.2ex}\color{black} 0.302$\pm${\scriptsize 0.134} & \cellcolor{gray!16}\rule{0pt}{2.2ex}\color{black} 0.395$\pm${\scriptsize 0.258} & \textbf{\cellcolor{gray!20}\rule{0pt}{2.2ex}\color{black} 0.513$\pm${\scriptsize 0.239}}$^{\dagger}$$^{*}$ & \rule{0pt}{2.2ex}0.009 & \rule{0pt}{2.2ex}0.039 \\
\bottomrule \\
\multicolumn{8}{r}{\footnotesize Significance marks indicate $p<0.05$: $^{*}$ for +Self-alignment vs PCGRLLM, $^{\dagger}$ for Zero-shot vs PCGRLLM.}
\end{tabular}
\end{table*}

\subsection{Hyperparameters}
\label{sec:pcg_agent_parameters}
The convergence timesteps for different reward functions were determined experimentally, and the detailed neural network architecture followed the design used in the original environment, as summarized in Table \ref{tab:Hyperparameter}.

\begin{table}[!h]
\centering
\caption{Hyperparameters and DRL agent network architecture used in the experiments.}
\label{tab:Hyperparameter}
\begin{tabular}{m{3.5cm}m{3.5cm}}
\toprule
\textbf{Parameter}      &   \textbf{Value}  \\
\midrule
\textbf{PCGRL Agent Setting}          &                   \\
Baseline              &   PPO            \\
$\lambda_{\text{GAE}}$              &   0.95            \\
Epochs size                    &   10              \\
Rollout length                      &   128             \\
Minibatch size                      &   4               \\
Clipping coefficient ($\epsilon$)   &   0.2             \\
Learning rate                       &   0.0001          \\
Value loss coefficient              &   0.5             \\
Entropy coefficient                 &   0.01            \\
Maximum gradient norm               &   0.5             \\
$\gamma$                            &   0.99            \\
Maximum steps                       &   50,000,000      \\
\midrule
\textbf{Network Architecture}       &                   \\
Conv layers &   [31, 31, 3] → [16, 15, 15] \\
            &   [16, 15, 15] → [8, 8, 8]   \\
Actor network &   [4096] → [64] → [2]         \\
Critic network &   [3844] → [64] → [1]         \\
\bottomrule
\end{tabular}
\end{table}

\subsection{Open-weight Model Performance}
\label{sec:open_weight_model_performance}
To verify the generalization of the proposed framework across different LLM backbones, we additionally conducted experiments with an open-weight model.  
Specifically, we employed Meta's \texttt{llama3.2-90b-instruct} \cite{touvron2023llama} available in Amazon Bedrock.
Table \ref{tab:performance_llama} presents the ablation results by separating the self-alignment mechanism and the feedback module.  
PCGRLLM achieved the highest score in most cases, showing that the proposed method generally improves performance even when applied to a different LLM backbone.  
\\
These improvements are statistically significant over zero-shot ($p (\text{ZS})<0.01$) and further significant over SA ($p(\text{ZS})<0.05$), particularly in the CoT and GoT settings.  
Overall, the results confirm that feedback not only mitigates the limitations of zero-shot generation but also provides clear benefits beyond the previously proposed self-alignment approach.  
In particular, feedback enables the correction of shortcomings in agents trained with imperfect reward functions, thereby leading to more reliable and effective policy learning.

\subsection{Full Result on Specificity of Feedback}
\label{sec:full_result_feedback}

Table~\ref{tab:specificity_feedback_full} reports the full iteration-wise results, revealing consistent trends across PE methods and feedback types. In particular, ToT and GoT benefited the most from structured feedback, with the use of Specific feedback leading to noticeable performance gains across successive iterations. By contrast, CoT exhibited only moderate sensitivity to feedback, reaching its highest scores around the third to fourth iteration before plateauing. These results suggest that structured feedback, especially when provided as specific guidance, plays a critical role in boosting both stability and peak performance compared to conditions without feedback.

\begin{table*}
\centering
\caption{Full results on performance across feedback types. Iteration-wise performance by feedback type and reasoning methods, with the best iteration in each row shown in bold.}
\label{tab:specificity_feedback_full}
\begin{tabular}{p{2.0cm}p{2.0cm} c c c c c c}
\toprule
Reasoning & Feedback & $y_{1}$ & $y_{2}$ & $y_{3}$ & $y_{4}$ & $y_{5}$ & $y_{6}$ \\
\midrule
CoT \cite{wei2022chain} & No & 0.293$\pm${\scriptsize 0.201} & 0.328$\pm${\scriptsize 0.227} & 0.244$\pm${\scriptsize 0.228} & \textbf{0.351}$\pm${\scriptsize 0.230} & 0.326$\pm${\scriptsize 0.215} & 0.309$\pm${\scriptsize 0.240} \\
CoT \cite{wei2022chain} & Generic & 0.344$\pm${\scriptsize 0.207} & \textbf{0.408}$\pm${\scriptsize 0.216} & 0.364$\pm${\scriptsize 0.267} & 0.391$\pm${\scriptsize 0.269} & 0.281$\pm${\scriptsize 0.216} & 0.365$\pm${\scriptsize 0.264} \\
CoT \cite{wei2022chain} & Specific & 0.162$\pm${\scriptsize 0.207} & 0.198$\pm${\scriptsize 0.227} & \textbf{0.361}$\pm${\scriptsize 0.268} & 0.203$\pm${\scriptsize 0.270} & 0.160$\pm${\scriptsize 0.123} & 0.284$\pm${\scriptsize 0.308} \\
\midrule
ToT \cite{yao2024tree} & No & 0.268$\pm${\scriptsize 0.216} & 0.268$\pm${\scriptsize 0.228} & 0.258$\pm${\scriptsize 0.161} & 0.241$\pm${\scriptsize 0.138} & \textbf{0.274}$\pm${\scriptsize 0.229} & 0.250$\pm${\scriptsize 0.169} \\
ToT \cite{yao2024tree} & Generic & 0.235$\pm${\scriptsize 0.177} & 0.264$\pm${\scriptsize 0.214} & \textbf{0.297}$\pm${\scriptsize 0.220} & 0.266$\pm${\scriptsize 0.190} & 0.257$\pm${\scriptsize 0.193} & 0.232$\pm${\scriptsize 0.170} \\
ToT \cite{yao2024tree} & Specific & 0.160$\pm${\scriptsize 0.255} & 0.185$\pm${\scriptsize 0.219} & 0.189$\pm${\scriptsize 0.176} & 0.339$\pm${\scriptsize 0.342} & 0.338$\pm${\scriptsize 0.367} & \textbf{0.360}$\pm${\scriptsize 0.313} \\
\midrule
GoT \cite{besta2024graph}  & No & \textbf{0.367}$\pm${\scriptsize 0.184} & 0.256$\pm${\scriptsize 0.219} & 0.282$\pm${\scriptsize 0.225} & 0.279$\pm${\scriptsize 0.231} & 0.296$\pm${\scriptsize 0.242} & 0.272$\pm${\scriptsize 0.247} \\
GoT \cite{besta2024graph} & Generic & 0.289$\pm${\scriptsize 0.218} & 0.309$\pm${\scriptsize 0.163} & 0.258$\pm${\scriptsize 0.190} & 0.345$\pm${\scriptsize 0.202} & 0.306$\pm${\scriptsize 0.180} & \textbf{0.385}$\pm${\scriptsize 0.184} \\
GoT \cite{besta2024graph} & Specific & 0.009$\pm${\scriptsize 0.022} & 0.327$\pm${\scriptsize 0.307} & 0.301$\pm${\scriptsize 0.329} & 0.185$\pm${\scriptsize 0.208} & 0.413$\pm${\scriptsize 0.358} & \textbf{0.423}$\pm${\scriptsize 0.329} \\
\bottomrule
\end{tabular}
\end{table*}

\subsection{Vision-based Feedback Analysis}
\label{sec:vision_feedback}

\begin{table}[!h]
\caption{Ablation study results for framework architecture, comparing accuracy across prompt engineering methods and scenarios.}
\label{tab:feedback_input}
\begin{tabular}{p{1.5cm}|p{0.73cm}p{0.73cm}p{0.73cm}|p{0.73cm}p{0.73cm}p{0.73cm}}
\toprule
\textbf{Feedback Input} & \multicolumn{3}{c|}{\textbf{Text}} & \multicolumn{3}{c}{\textbf{Image}} \\
 & ZS & +FB & $\Delta{}$ & ZS & +FB & $\Delta{}$ \\
PE &  &  &  &  &  &  \\
\midrule
CoT & 0.033 & 0.156 & +0.122 & 0.031 & 0.157 & +0.126 \\
\bottomrule
\end{tabular}
\end{table}

\begin{figure}[!h]
    \centering
    \begin{tabular}{cc} 
        \includegraphics[width=0.20\textwidth]{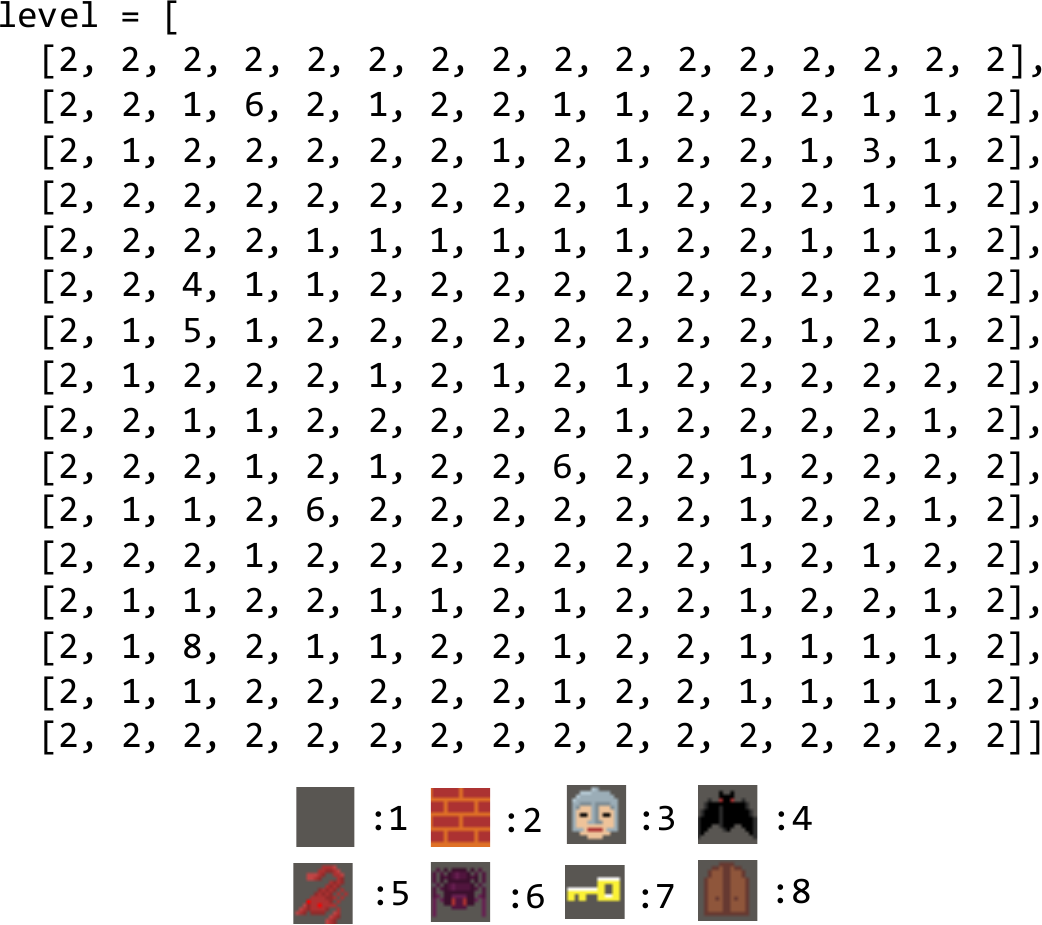} & 
        \includegraphics[width=0.20\textwidth]{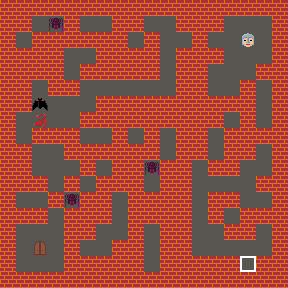} \\
        (a) Text Input & (b) Image Input
    \end{tabular}
    \caption{Comparison of input methods: (a) textual input, where the 2D array is represented as plain text, and (b) image input, where the array is converted into an image format for processing.}
    \label{fig:feedback_input_example}
\end{figure}

The state-of-the-art Vision-Language Models (VLMs), such as \texttt{gpt-4o} with vision capabilities, can process visual inputs to perform tasks like question answering based on a given image.
This section investigates the reasoning capabilities of VLMs, specifically evaluating whether they have been trained on and can effectively reason about the distribution of game-rendered images.
Fig. \ref{fig:feedback_input_example} illustrates the difference between two input methods: textual input, where a 2D array is represented as plain text, and image input, where the same array is processed in visual form.
We evaluated how the modality of information provided when generating feedback influences changes in accuracy performance.

Table \ref{tab:feedback_input} provides quantitative evidence of this improvement. In a CoT reasoning, both textual and image inputs show performance gains with the addition of feedback (+FB). For textual input, the accuracy improves from 0.033 to 0.156, marking a gain of +0.122. Similarly, for image input, the accuracy increases from 0.031 to 0.157, with a comparable gain of +0.126. These results suggest that feedback is highly effective across both modalities, leading to substantial performance enhancements. Moreover, the comparable improvements indicate that both text and image inputs benefit similarly from feedback, emphasizing the importance of leveraging feedback mechanisms in multimodal frameworks to align reward functions with complex, goal-oriented tasks.
This also highlights the potential for evaluating and providing feedback on high-dimensional outputs, such as gameplay videos, to improve the quality of generated content in complex scenarios.

\subsection{Codebase and Prompts}
\label{sec:codebase}

The full implementation and experimental resources are available at \href{https://github.com/bic4907/pcgrl-llm}{https://github.com/bic4907/pcgrl-llm}. 
The repository contains source code for our level generation framework, evaluation scripts, and prompt templates. 
In addition, detailed instructions for the self-alignment and feedback mechanisms are provided in the supplementary material.

\subsection{Comparative Analysis of Human and LLM Reward Design}
\label{sec:human_crafted_reward}
Table \ref{tab:performance_pcgrllm} shows that the relative effectiveness of human-crafted versus LLM-generated rewards is scenario-dependent, suggesting a complementary division of roles in reward design. In Scenarios 1 and 2, which involve multiple conditional objectives, the human baseline performs better, likely because human-designed rewards leverage domain expertise to combine multiple reward signals and finely calibrate trade-offs among competing objectives. In contrast, in Scenarios 3 and 4—dominated by spatial constraints—the LLMs-generated rewards outperform the human baseline, indicating that LLMs can more readily construct effective reward formulations that capture large-scale spatial structure.

Overall, these findings suggest a complementary human–LLM workflow for reward design. LLMs are effective at synthesizing coarse reward functions that encode spatial structure, whereas human experts are better suited to refining these functions by tuning parameters and coordinating multiple interacting conditions. This division of labor implies that LLM-generated rewards can serve as strong initial formulations, which can then be improved through human-guided adjustments to satisfy nuanced multi-objective requirements and avoid subtle failure modes.

\onecolumn
\section*{Supplementary Material}
\label{sec:supple}

\subsection*{A. Natural Language Prompt}
\label{sec:prompt_example}

\begin{tcolorbox}[colback=white, colframe=gray, title=Reward Refinement, breakable]

{\textbf{\MakeUppercase{Instruction}}}  \\

\textbf{PCG Agent Reward Function Generation Task}

You are a reward function engineer trying to write reward functions to solve reinforcement learning tasks as effective as possible.\\

\textbf{PCGRL Environment}

The gym-pcgrl project provides an environment for Procedural Content Generation via Reinforcement Learning (PCGRL), where agents learn to generate game levels.
One of the environments, "binary," involves creating maps using given tile types.
In the "narrow" representation, the agent modifies the map by scanning through it, changing tiles one by one.
This localized approach encourages more precise level design.
The agent receives rewards based on how well the generated level meets the predefined objectives.
If the agent generates a level that aligns closely with the goal—such as creating a functional and balanced map—it receives a higher reward.
This reward-driven mechanism incentivizes the agent to improve its design strategies over time, learning to create optimal levels through trial and error.\\

\textbf{Reward Function}

The reward function is a function that calculates the reward value for the agent based on the playtested results.
The function is written in Python and loads the playtested results from the json file and calculates the reward value based on the results.

[colback=white, colframe=gray, title=Reward Function Guidelines, breakable]
\begin{lstlisting}[language=Python]
import jax.numpy as jnp

def compute_reward(prev_array, prev_stats, curr_array, curr_stats) -> float:
    reward = 0.0

    return reward
\end{lstlisting}

'prev\_array' and 'curr\_array' are two-dimensional array representing the game level with tile numbers. 
The range of width is different by the array input (e.g., 10-20), so write a function works on any width and height.

\textbf{prev\_array}:  previous game level represented with tile numbers (jnp.array, (h,w))

\textbf{curr\_array}: current game level represented with tile numbers (jnp.array, (h,w))\\
The array is a 2D array with the shape of (height, width) to represent the game level.\\

The level is represented with tile numbers. The tile number is an integer value.

\textbf{Tile Number}

EMPTY = 1, WALL = 2, PLAYER = 3, BAT = 4, SCORPION = 5, SPIDER = 6, KEY = 7, DOOR = 8\\

\textbf{Thought Tips}

You have only one chance to generate a reward function. Make a reward function to work on the environment \\

\textbf{Target Scenario}

The 'Player' needs to obtain the Key and escape through the 'Door'. To pick up the key, the player must encounter one of the monsters: BAT. The players can figure out all of the monsters when they play the level several times.

Ensure that the generated map includes a reasonable number of monsters and keys.
First, create a reward function that minimizes the generation of monsters other than those mentioned above.
Second, design the reward function so that the map does not contain an excessive number of monsters or keys.\\

\end{tcolorbox}

\vspace{1em}

\begin{tcolorbox}[colback=white, colframe=gray, title=Self-Alignment, breakable]
{\textbf{\MakeUppercase{Instruction}}}  \\

\textbf{PCG Agent Reward Function Generation Task}

Here is your generated reward function and the result of the code execution (Code Result) with random action agent.
Keep the existing function structure and tune small components like weights to adjust the reward values, ensuring the logic remains intact. 

Focus on reward scale and sparsity.  

Do not hallucinate reward function.\\

\textbf{Reward Function Code}

\{previous reward function\} \\

Code Result

Mean: -646.652710

Std: 502.900909

Zero Value Percent: 23.9173\% 

\medskip
\noindent\textcolor{gray}{\rule{\linewidth}{0.4pt}}

{\textbf{\MakeUppercase{Response}}}  \\

\textbf{Revised Reward Function Code}

\{Revised reward function\} 
\\

\textbf{Changes Made}

- Increased the reward for bats and scorpions to 5.0 to encourage their presence more strongly.

- Increased the penalty for spiders to 7.0 to discourage their presence more strongly.

- Increased the reward for keys to 4.0 to encourage collecting keys.

- Increased the penalty for having more than 2 keys to 4.0 to discourage excessive key collection.

- Increased the penalty for having more than 3 total monsters to 6.0 to discourage excessive monster presence.

\end{tcolorbox}

\vspace{1em}

\begin{tcolorbox}[colback=white, colframe=gray, title=Feedback, breakable]
{\textbf{\MakeUppercase{Instruction}}}  \\

\textbf{Evaluate why the current reward function is producing these game levels}

Analyze how the reward structure influences the design decisions made by the AI.

For each of the following aspects, explain potential issues and how modifying the reward system could lead to better outcomes. \\

\textbf{The Following Content was Generated Based on the Current Reward Function}

Review the generated levels and identify common problems in the contents.\\

\textbf{Previous Reward Code}

\{previous reward code\} \\

\textbf{Evaluating the Generated Levels: Step-by-Step Analysis}

1. Review multiple generated levels and pinpoint common problems in the contents.

2. Decide on one specific problem and analyze why the current reward function causes this issue.

- If PCGRL fails to train, it could indicate a sparse reward problem. Consider the reward value and frequency in the previous implementation.

- Another issue could be overfitting on specific actions, such as repeatedly placing a tile excessively, leading to excessive rewards for certain actions.

- Since predicting reward values from a complex function can be difficult, a simpler reward function might resolve training problems.

3. Pinpoint problematic parts in the reward function and propose changes. Provide the changed part only of the reward function.

\medskip
\noindent\textcolor{gray}{\rule{\linewidth}{0.4pt}}

{\textbf{\MakeUppercase{Response}}}  \\

\textbf{Feedback: Step-by-Step Analysis}

\textbf{1. Review of Generated Levels and Common Problems:}

- Imbalance in Monster Types:

Across the generated levels, a common issue is the lack of a balanced distribution of monster types. Some levels have an overrepresentation of one type of monster, while others lack diversity in monster types.

- Key and Door Placement Issues:

Another issue is the placement of keys and doors. In some levels, keys are placed in locations that are not easily accessible or logical in terms of gameplay flow. \\

\textbf{2. Analysis of the Current Reward Function:}

- Lack of Incentive for Balance:

The current reward function provides a reward for having at least one of each required monster type (BAT, SCORPION, SPIDER), but it does not sufficiently incentivize a balanced distribution of these monsters. The penalty for imbalance is relatively small, which may not be enough to encourage diversity.

[...omitted for brevity] \\

\textbf{Revised Reward Function (Changed Parts Only):}

\begin{lstlisting}[language=Python]
# Increase penalty for imbalance in monster types
monster_balance_penalty = (
    jnp.abs(curr_bats - curr_scorpions)
    + jnp.abs(curr_scorpions - curr_spiders)
    + jnp.abs(curr_spiders - curr_bats)
)
reward -= monster_balance_penalty * 0.3  # Increased from 0.1 to 0.3

# Introduce reward for strategic placement of keys and doors
# (This is a conceptual addition; actual implementation would require additional logic to evaluate placement)
strategic_placement_reward = evaluate_placement(curr_array)  # Hypothetical function to evaluate placement
reward += strategic_placement_reward

\end{lstlisting}
\end{tcolorbox}

\vspace{1em}

\begin{tcolorbox}[colback=white, colframe=gray, title=Generated Reward Function Example, breakable]
\begin{lstlisting}[language=Python]
import jax.numpy as jnp

def compute_reward(prev_array, unused3, curr_array, unused4) -> float:
    # Define tile numbers
    EMPTY = 1
    WALL = 2
    PLAYER = 3
    BAT = 4
    SCORPION = 5
    SPIDER = 6
    KEY = 7
    DOOR = 8

    # Initialize reward
    reward = 0.0

    # Count the number of each tile type in the current and previous arrays
    def count_tiles(array, tile):
        return jnp.sum(array == tile)

    # Calculate the number of each relevant tile in the current and previous arrays
    curr_player_count = count_tiles(curr_array, PLAYER)
    curr_key_count = count_tiles(curr_array, KEY)
    curr_door_count = count_tiles(curr_array, DOOR)
    curr_bat_count = count_tiles(curr_array, BAT)
    curr_scorpion_count = count_tiles(curr_array, SCORPION)
    curr_spider_count = count_tiles(curr_array, SPIDER)

    prev_player_count = count_tiles(prev_array, PLAYER)
    prev_key_count = count_tiles(prev_array, KEY)
    prev_door_count = count_tiles(prev_array, DOOR)
    prev_bat_count = count_tiles(prev_array, BAT)
    prev_scorpion_count = count_tiles(prev_array, SCORPION)
    prev_spider_count = count_tiles(prev_array, SPIDER)

    # Reward for having exactly one player, one key, and one door
    reward += 10.0 * (curr_player_count == 1) - 10.0 * (prev_player_count == 1)
    reward += 10.0 * (curr_key_count == 1) - 10.0 * (prev_key_count == 1)
    reward += 10.0 * (curr_door_count == 1) - 10.0 * (prev_door_count == 1)

    # Reward for having at least one of each monster type
    reward += 5.0 * (curr_bat_count >= 1) - 5.0 * (prev_bat_count >= 1)
    reward += 5.0 * (curr_scorpion_count >= 1) - 5.0 * (prev_scorpion_count >= 1)
    reward += 5.0 * (curr_spider_count >= 1) - 5.0 * (prev_spider_count >= 1)

    # Penalize more heavily for having more than 3 of any monster type
    reward -= 15.0 * (curr_bat_count > 3) - 15.0 * (prev_bat_count > 3)
    reward -= 15.0 * (curr_scorpion_count > 3) - 15.0 * (prev_scorpion_count > 3)
    reward -= 15.0 * (curr_spider_count > 3) - 15.0 * (prev_spider_count > 3)

    # Penalize for having more than 1 key
    reward -= 10.0 * (curr_key_count > 1) - 10.0 * (prev_key_count > 1)

    # Reward for balanced distribution of monster types
    balanced_monsters = (curr_bat_count <= 3) & (curr_scorpion_count <= 3) & (curr_spider_count <= 3)
    reward += 10.0 * balanced_monsters.astype(float)

    # Calculate distances between important tiles
    def calculate_distance(array, tile1, tile2):
        positions1 = jnp.nonzero(array == tile1, size=1, fill_value=-1)
        positions2 = jnp.nonzero(array == tile2, size=1, fill_value=-1)
        pos1 = jnp.array([positions1[0][0], positions1[1][0]])
        pos2 = jnp.array([positions2[0][0], positions2[1][0]])
        return jnp.linalg.norm(pos1 - pos2)

    # Reward for proximity of PLAYER, KEY, and DOOR
    threshold = 5.0
    player_key_distance = calculate_distance(curr_array, PLAYER, KEY)
    key_door_distance = calculate_distance(curr_array, KEY, DOOR)
    reward += 5.0 * (player_key_distance < threshold) + 5.0 * (key_door_distance < threshold)

    return reward

\end{lstlisting}
\end{tcolorbox}

\end{document}